\documentclass[useAMS,usenatbib]{tellus}
\usepackage[a4]{crop}
\usepackage{amsmath}
\usepackage{graphicx}
\usepackage{soul}
\usepackage{fancyhdr}
\usepackage{appendix}

\usepackage[english]{babel}
\usepackage{enumitem}
\usepackage[letterpaper,top=2cm,bottom=2cm,left=3cm,right=3cm,marginparwidth=1.75cm]{geometry}

\usepackage{natbib}
\usepackage{multirow} 
\usepackage{amsmath}
\usepackage{amssymb}
\usepackage{amsfonts}
\DeclareMathOperator*{\argmax}{arg\,max}
\DeclareMathOperator*{\argmin}{arg\,min}
\usepackage{graphicx}
\usepackage[colorlinks=true, allcolors=blue]{hyperref}
\usepackage{booktabs}
\usepackage{algorithm}
\usepackage{algorithmic} 
\usepackage{todonotes}

\newcommand{\propel}{{\sc propel}}
\newcommand{\prop}{{\sc prop}}
\newcommand{\bd}{\mathbf{d}}
\newcommand{\app}{\tilde{{\cal O}}}
\newcommand{\opt}{{\cal O}}

\title{\propel{}: Supervised and Reinforcement Learning for Large-Scale Supply Chain Planning}
\author{Vahid Eghbal Akhlaghi, Reza Zandehshahvar, and Pascal Van Hentenryck \\}
\affiliation{NSF Artificial Intelligence Institute for Advances in Optimization (AI4OPT) \\
  Georgia Institute of Technology \\
  Email: pvh@gatech.edu}

\begin{document}
\maketitle

\begin{abstract}
This paper considers how to fuse Machine Learning (ML) and
optimization to solve large-scale Supply Chain Planning (SCP) optimization
problems.  These problems can be formulated as MIP models which
feature both integer (non-binary) and continuous variables, as well as
flow balance and capacity constraints. This raises fundamental
challenges for existing integrations of ML and optimization that have
focused on binary MIPs and graph problems. To address these, the paper
proposes \propel{}, a new framework that combines optimization with
both supervised and Deep Reinforcement Learning (DRL) to reduce the
size of search space significantly. \propel{} uses supervised
learning, not to predict the values of all integer variables, but to
identify the variables that are fixed to zero in the optimal
solution, leveraging the structure of SCP applications.  \propel{}
includes a DRL component that selects which fixed-at-zero variables must
be relaxed to improve solution quality when the supervised learning step
does not produce a solution with the desired optimality
tolerance. \propel{} has been applied to industrial supply chain
planning optimizations with millions of variables. The computational
results show dramatic improvements in solution times and quality,
including a 60\% reduction in primal integral and an 88\% primal gap
reduction, and improvement factors of up to 13.57 and 15.92,
respectively. \\
\end{abstract}

\section{Introduction}\label{intro}

This paper considers the fusion of machine learning and optimization
for finding near-optimal solutions to large-scale Supply Chain
Planning (SCP) applications in reasonable times. These optimizations are
often expressed as Mixed Integer Linear Programs (MIPs) of the form
\begin{equation*}
\begin{aligned}
& \phi(\mathbf{d}) = & & \argmax_{\textbf{x}} & &  \textbf{c}^T \textbf{x} \\
& & &  \text{subject to} & & \textbf{Ax} \leq \textbf{b} \\
  & & & & & \textbf{x} \in \mathbb{Z}^q \times \mathbb{R}^{n-q} 
\end{aligned}
\end{equation*}
where $\textbf{x} = (x_1, \ldots, x_n)^T$ comprises $q$ integer
variables and $n-q$ continuous variables, and $\mathbf{d} =
(\textbf{A}, \textbf{b}, \textbf{c})$ denote the MIP instance data. In
these applications, integer variables represent production decisions
and inventory levels that span a wide range of possible
values. Moreover, these optimizations typically involve millions of
variables, making them particularly hard to solve within the time
requirements of practical settings (e.g., with planners in the loop).

Fortunately, in practical applications, the same optimization problem
is solved repeatedly with different instance data. For instance, in
supply chain planning optimization, the demand parameter ${\bf d}$
varies over time and is an exogenous random variable. Moreover,
planners typically work with multiple demand forecast scenarios to
understand possible outcomes, hedge against uncertainty, assess the
resilience of their operations and, more generally, manage financial
risks. Planners thus face what is often called {\em parametric
  optimization problems}, i.e., applications that share the same
overall structure and must be solved for numerous instance data that
are typically related. This presents an opportunity of {\em learning
  to solve} such a parametric optimization offline instead treating
each instance as a new optimization task each time. By shifting most
of the computational burden offline, the hope is that these instances
can be solved an order of magnitude faster and meet the expectations
of supply chain planners.

The use of machine learning for Combinatorial Optimization (CO) has
attracted significant attention in recent years and is increasingly
recognized a promising direction to overcome some of the computational
challenges \citep{bengio2021machine,kotary2021end,
  vesselinova2020learning}. Learning-based approaches are well-known
for their capability to yield effective empirical algorithms,
leveraging regularities in real-world large datasets
\citep{li2018combinatorial, khalil2022mip}, and improving solution
times \citep{cappart2021combining,kotary2021end}. As described in
Section \ref{section:related}, much of the research in learning to
optimize combinatorial problems has focused on graph problems (or
problems that are naturally expressed as graph problems) and binary
MIPs. Supply chain planning optimizations are fundamentally different
in nature: (1) they are expressed in terms of integer variables that
can take large values, as they represent procurement, production, and
inventory decisions over a long time horizon; and (2) they are
typically large-scale, featuring millions of integer variables. As a
result, many existing techniques that fuse machine learning and
optimization are not directly applicable. In particular, the
satisfaction of flow and inventory constraints is
particularly challenging.

This paper is motivated by the need to address these challenges and
progress supply chain planning optimization. It introduces \propel{},
a novel framework that combines supervised and deep reinforcement
learning to find near-optimal solutions to large-scale industrial
problems. \propel{} uses supervised learning, not to predict the
values of all integer variables, but to identify the variables that
are fixed to zero in the optimal solution. By not assigning non-zero
variables, \propel{} avoids generating infeasible solutions that may
be hard to repair.  Given the nature of supply chain planning
optimization, many variables are fixed at zero in optimal solutions
and hence the supervised learning phase leads to significant
reductions in the number of variables and solution time of the
resulting optimization. However, in some cases, the solution quality
may not be within the desired optimality tolerance. To remedy this
limitation, \propel{} includes a Deep Reinforcement Learning (DRL)
component that selects which fixed-at-zero variables must be relaxed
(unfixed).  \propel{} has been applied on industrial supply chain
planning optimizations with millions of variables. The computational
results show dramatic improvements in solution times and quality,
including a 60\% reduction in primal integral and an 88\% primal gap
reduction. They also show the benefits of the DRL component, which
improves the primal gap by a significant factor on the most challenging
instances. 

The main contributions of this paper, and \propel{} in particular, can
be summarized as follows.
\begin{itemize} 
\item To the best of our knowledge, \propel{} is the first framework
  that fuses ML and optimization for industrial supply chain planning
  optimizatiom, filling a gap identified in the literature
  \citep{tirkolaee2021application}.

\item \propel{} combines the benefits of supervised and deep
  reinforcement learning. Its supervised component does not attempt to
  predict the values of all variables, only those which are fixed at
  zero. This makes it possible to handle complex
  constraints. Moreover, its DRL component enables \propel{} to
  overcome prediction errors by relaxing some of the fixed-at-zero
  variables. The supervised learning module of \propel{} leverages the
  concept of reduced costs to minimize wrong predictions.

\item \propel{} was rigorously tested using realistic, large-scale
  supply chain instances. It thus addresses a common criticism of
  learning to optimize research, i.e., the reliance on simulated or
  small-scale data to demonstrate effectiveness \citep{ni2020systematic}.

\item \propel{} provides dramatic improvements in primal integral and
  primal gap metrics compared to state-of-the-art MIP solvers.
\end{itemize}

\noindent
The paper is organized as follows. Section \ref{section:related}
presents the related work, Section \ref{section:methodology} gives an
overview of \propel{}, and Section \ref{section:learning} specifies
the learning task. Sections \ref{section:supervised} and
\ref{section:DRL} are the core of the paper: they present the
supervised and DRL components. Section \ref{section:SCP} gives a
stylized presentation of supply chain planning optimization. Section
\ref{section:results} presents the computational results and Section
\ref{section:conclusion} concludes the paper.

\section{Literature Review}
\label{section:related}
The integration of optimization and machine learning recently led to
the development of several distinct approaches, surveyed
by \cite{kotary2021end}. These methods, often referred to by various
names in the literature, can be categorized into three main categories
(which are increasingly hybridized). {\em Decision-focused learning}
(also called Smart-Predict-Then-Optimize)
(e.g., \citep{elmachtoub2022smart,el2019generalization,ning2019optimization,sahinidis2004optimization,donti2017task})
aims at training forecasting and optimization models in the same
pipeline. {\em Optimization proxies}
(e.g., \citep{chen2023end,kotary2021end,kotary2022fast,donti2021dc3,
huang2021deepopf, park2023self}) aims at learning the mapping between
the input and an optimal solution of an optimization problem. {\em
Learning to optimize} is concerned with a variety of techniques to
improve the efficiency and the solution quality of optimization
algorithms and solvers. This paper is concerned with the third
category.

Optimization proxies have accumulated great success for continuous
optimization problems, but they encounter feasibility and training
challenges in the presence of discrete variables
(see \cite{tran2021differentially, fioretto2020predicting,
detassis2021teaching}). Learning models for discrete optimization
problems lack useful gradients, as the arg-max operator in discrete
problems is piecewise constant, complicating backpropagation. One way
to address this, is to construct effective approximations of the
gradients, e.g., by generating continuous surrogates of the MIP to
facilitate effective training (see \cite{ferber2020mipaal,
kotary2021end, donti2017task,wilder2019melding}). As an
alternative, \cite{park2023confidence} introduced the
\textit{Predict-Repair-Optimize} framework,
which predicts an optimal solution, fixes a subset of variables with
confident predictions, and restores feasibility with a dedicated
repair algorithm. A final optimization step is applied to complete the
partial assignment. \propel{} drew inspiration from this framework,
but differs in two key aspects: (1) only fixing variables to zero and
leaving non-zero integer variables free; and (2) using deep
reinforcement learning to determine which variables to unfix in order
to obtain the desired optimality target. \propel{} also features some
important innovations in its supervised learning models.


Learning to optimize encompasses a wealth of approaches, many of which
are reviewed
in \citep{lodi2017learning,bengio2021machine,kotary2021end}. They
include techniques to guide search decisions in branch and bound/cut
solvers \citep{zarpellon2021parameterizing,gasse2019exact,gupta2020hybrid,tang2020reinforcement}
and direct the application of primal heuristics within
branch-and-bound \citep{khalil2017learning,chmiela2021learning,bengio2020learning,song2020general}. While
it differs from these studies, \propel{} shares some similarities with
those focusing on variable selection
strategies \citep{khalil2016learning,alvarez2017machine,balcan2018learning}. \cite{khalil2016learning}
highlight that, beyond the supervised learning approaches prevalent in
this context, reinforcement learning formulations are worth exploring
due to the sequential nature of the variable selection task. Aligned
with this observation, \propel{} implements a novel approach that
combines the benefits of supervised learning and deep reinforcement
learning.

To tackle combinatorial optimization problems in reinforcement
learning pipelines, \cite{bello2016neural} utilize reinforcement
learning to train \textit{pointer networks} for solving synthetic
instances of the planar traveling salesman problem (TSP) with up to
100 nodes, and demonstrated their approach on synthetic random
Knapsack problems with up to 200 elements. Pointer networks, initially
introduced by \cite{vinyals2015pointer}, employ an architecture where
an encoder, typically a \textit{recurrent neural network (RNN)},
processes all nodes of an input graph to generate node encodings, uses
a decoder, also an RNN, which leverages an attention mechanism akin
to \cite{bahdanau2014neural} to produce a probability distribution
across over the encoded nodes. By iterating this decoding process, the
network can generate a permutation of the input nodes, effectively
solving permutation-based optimization problems. These models are
typically trained via supervised learning using precomputed solutions
of small-scale planar TSP instances (up to 50 nodes) as
targets. Pointer networks and their variants with RNN decoders are
specialized in solving problems like TSP and Vehicle Routing Problem
(VRP) \citep{vinyals2015pointer, kool2018attention,
nazari2018reinforcement}. In contrast, \cite{kool2018attention}
introduced a variant that incorporates prior knowledge using a graph
neural network (GNN) instead of an RNN decoder. This adaptation aims
at achieving input node order invariance, thereby improving learning
efficiency and computational performance.

The burgeoning role of GNNs in bridging ML with combinatorial
optimization is reviewed by \cite{cappart2021combining}. Many problems
are inherently graph-based, either through direct representation
(e.g., routing on road networks) or by representing
variable-constraint interactions in MIP models as bipartite
graphs \citep{khalil2022mip}. \propel{} is capable of using this
latter method for feature extraction, as detailed in Appendix
A. \cite{ding2020accelerating} pioneered the use of GNNs on a
tripartite graph comprising variables, constraints, and a unique
objective node, aiming at identifying \textit{stable variables}
consistent across various solution sets, thereby aiding the pursuit of
optimal solutions through learned patterns. However, the existence of
such stable variables may not be consistent across all combinatorial
optimization problems. Concurrently, \cite{khalil2022mip} introduced
the MIP-GNN framework, which shifts the focus to predicting the
likelihood of binary variables in near-optimal solutions by encoding
variable-constraint interactions as a bipartite graph without the
objective node. This approach seeks to enhance the heuristic
components of problem-solving methods. This approach applies
to \textit{Binary} MIPs that typically feature up to only tens of
thousands of variables and constraints. \propel{} differs from these
methods in several ways: (1) it considers problems with arbitrary
integer; (2) it has been applied to applications with millions of
decision variables; and (3) it is not restricted to graph-based
problems but applies to standard MIP formulations.

\cite{li2018combinatorial} employ a \textit{graph convolutional
network (GCN)} to predict a set of probability maps for decision
variables, encoding the likelihood of each vertex being in the optimal
solution. Their approach enhances a problem-specific tree search
algorithm with ML and has shown promising results, although it is
specialized for node elimination on graphs and is not applicable to
general problems. \cite{nair2020solving} propose a more general
approach that first learns the conditional distribution in the
solution space via a GNN and then fixes a subset of discrete
variables, simplifying the MIP problem for computational efficiency.
However, fixing discrete variables may render the MIP model
infeasible. \cite{han2023gnn} introduce a
novel \textit{Predict-then-Search} framework that adopts the trust
region method, which searches for near-optimal solutions within a
well-defined region. Their approach utilizes a trained GNN model to
predict marginal probabilities of \textit{binary} variables in a given
MIP instance and subsequently searches for near-optimal solutions
within the trust region based on these
predictions. \cite{huang2024contrastive} propose the ConPaS framework
that learns to predict solutions to MILPs with contrastive
learning. ConPaS collect both high-quality solutions as positive
samples and low-quality or infeasible solutions as negative samples,
and learn to make discriminative predictions by contrasting the
positive and negative samples. It then fixes the assignments for a
subset of integer variables and then solves the reduced MIP to find
high-quality solutions. ConPaS was evaluated on four classes of binary
MIPs. Observe that these approaches predominantly test the performance
of their algorithms on binary problems. {\em One of the main
contributions of this paper is to show that \propel{} is effective in
producing near-optimal solutions to real-world non-binary MIPs.}

It is useful to position the contributions of \propel{} along several
axes to highlight some of its benefits.

\paragraph*{Utilization of Real Data}
\cite{ni2020systematic} highlight that currently, about
half of the data employed in SCP research originates from simulations
rather than real-world scenarios, with most research focusing on
mathematical models. This trend exists because analyzing historical
SCP data is challenging due to its complexity and scale. Consequently,
much of the research depends on artificially generated data, which
often fails to reflect the variability inherent in actual
operations. In contrast, \propel{} has been evaluated using
large-scale industrial case studies.

\paragraph*{Scalability}
In today's complex and dynamic world, as the volume of data increases,
the efficiency and effectiveness of traditional methods have
diminished \citep{tirkolaee2021application}. For instance, studies
addressing the TSP via ML struggle with performance degradation as
instance sizes exceed those encountered during
training \citep{bello2016neural, khalil2016learning,
kool2018attention, vinyals2015pointer}, confirming the challenges of
scaling to larger problems \citep{bengio2021machine}. Similarly, there
exists a significant disparity between the scale of actual SC networks
and the smaller academic test systems typically employed in
research. Most papers report numerical results on these smaller
systems, which are several orders of magnitude less complex than
real-world SCP applications. This discrepancy is particularly
concerning, as higher-dimensional data may adversely impact the
convergence and accuracy of machine learning algorithms according
to \cite{ni2020systematic}.  The experiments in
Section \ref{section:results} evaluates \propel{} on MIPs with
millions of variables and constraints, addressing problem sizes that
are significantly larger than those considered in the existing
literature.

\paragraph*{Transferability}
Many solution-generation methods for CO typically focus on problems
with specific solution structures and depend on strong assumptions to
develop their methods, limiting their applicability to certain problem
settings and policies.  For instance, approaches like Pointer
Networks \citep{vinyals2015pointer} and the Sinkhorn
layer \citep{emami2018learning} are predominantly suited for
sequence-based solution encoding, used to make a network output a
permutation—a constraint readily addressed by traditional CO
heuristics. However, many MIP problems, including the SCP
optimization studied in this paper, do not fit this permutation-based
representation. When the underlying assumptions or specific settings
fail to hold, the applicability and validity of the solution method
and results may be compromised.  Therefore, ensuring the
transferability of solution methods and results is crucial for broader
applicability and reproducibility in CO
research \citep{farazi2021deep}.  In the MIP context, approaches,
such as those by \cite{han2023gnn,khalil2022mip}, considered binary
problems. \propel{} does not have those restrictions and broadens the
class of MIPs that can benefit from ML. Its underlying techniques
are also rather general and may apply to other classes of
applications, a topic for future research.

%
%
%
%

\paragraph*{Feasible and High-Quality Solutions}

For certain classes of MIPs, finding feasible solutions or near
optimal solutions may be challenging, an issue that can be further
amplified when using machine learning methods
(e.g., \citep{pan2021deepopf, chen2022learning, nair2020solving,
yoon2022confidence}). For instance, in \textit{binary} MIPs, a
typical strategy is to train an NN to output a probability map in $[0,
1]^N$, where $N$ denotes the number of binary variables, indicating
the likelihood of each variable being one at optimality. Converting
such probability maps into discrete assignments frequently results in
infeasible solutions \citep{li2018combinatorial}. The method proposed
by \cite{han2023gnn} enhances feasibility and near-optimality by
exploring solutions within a trusted region around predicted points
rather than imposing fixed values.  \propel{} takes a different
approach: It initially reduces the size of the search space by fixing
variables to zero and avoiding fixing non-zero variables, and then
uses DRL to decide which variables to unfix, potentially enhancing
feasibility and near-optimality.

\section{Overview of \propel{}}
\label{section:methodology}
\begin{figure*}
\centering
\includegraphics[width=1\textwidth]{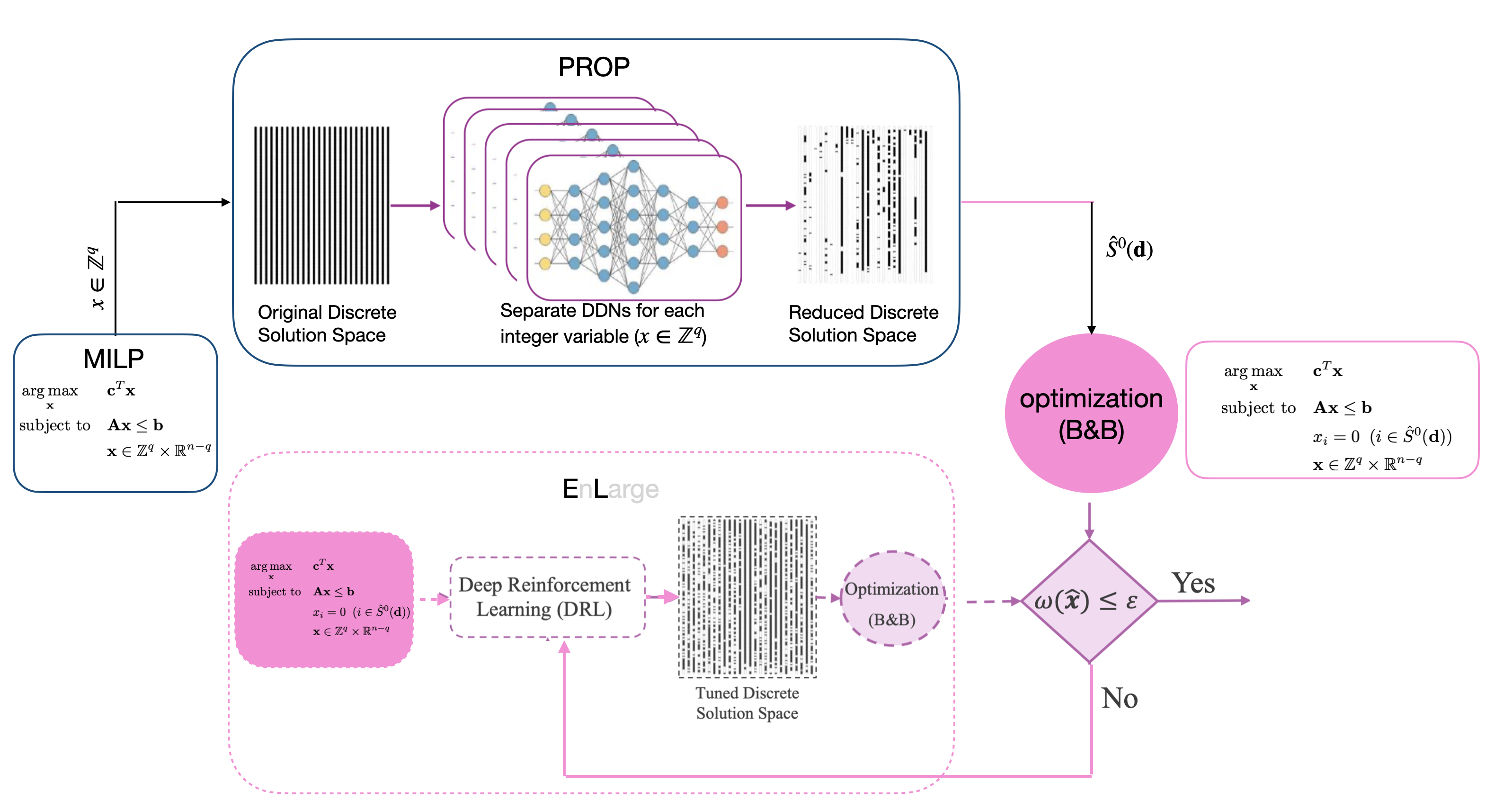}
\caption{An Overview of the \propel{} Framework.}
\label{fig:propel-overview}
\end{figure*}

\propel{} is a learning-based optimization framework designed to
improve the computational efficiency of large-scale industrial
MIPs. The architecture of \propel{} is shown in Figure
\ref{fig:propel-overview}. In a first phase (\prop{}), \propel{}
uses supervised learning to identify which integer variables are
likely to be fixed to zero in an optimal solution. The supervised
learning phase includes a key novelty: the use of reduced cost to
guide the selection of variables to be fixed at zero.  In a second
phase ({\sc EnLarge}), \propel{} uses deep reinforcement learning to
choose which fixed variables to relax when the optimality gap is not
within the desired tolerance. This second phase exploits temporal
features to partition the set of fixed variables and learn the
value-action function associated with their ``relaxations''.

\propel{} is motivated by the observation that, in many industrial
MIP problems from supply chain optimization, integer variables
represent production or inventory levels that span a wide range of
possible values. Given that there are multiple production pathways and
times to produce, a large number of integer variables are set to zero
at optimality. Predicting whether a variable is zero or non-zero,
rather than predicting its exact value, is advantageous in this
setting. By correctly identifying and fixing these zero-valued
variables, \propel{} simplifies the optimization task, reduces
unnecessary branching decisions, and improves the efficiency of
finding high-quality solutions. \propel{} also avoids making mistakes
in predicting integer values that can span a wide range of
values. Observe that predicting whether a variable is fixed at zero at
optimality is a classification problem, while predicting an exact
value is a regression task which will not output feasible solutions in
presence of constraints. Indeed, these predictions will often violate
capacity constraints and almost always fail to satisfy
inventory/balance equations, where integer variables interact with
multiple other variables. In many cases, however, such as those
discussed in Section \ref{feasAssure}, predicting whether a variable
is zero does not lead to infeasibilities, making \propel{} attractive
for applications in supply chain planning optimization.

\section{The Learning Task}
\label{section:learning}
Let $\mathbf{d} = (\textbf{A}, \textbf{b}, \textbf{c})$ denote the MIP
instance data. This paper considers a parametric MIP model
$\phi(\mathbf{d})$ defined as
\begin{equation}
\begin{aligned}
& \phi(\mathbf{d}) = & & \argmax_{\textbf{x}} & &  \textbf{c}^T \textbf{x} \\
& & &  \text{subject to} & & \textbf{Ax} \leq \textbf{b} \\
  & & & & & \textbf{x} \in \mathbb{Z}^q \times \mathbb{R}^{n-q} \label{MILP}
\end{aligned}
\end{equation}
where $\textbf{x} = (x_1, \ldots, x_n)^T$ contains $q$ integer
variables and $n-q$ continuous variables. The learning task aims at
simplifying the MIP model by predicting which integer variables are
fixed at zero in an optimal solution. Consider the set
\begin{equation}
  S^0(\mathbf{d}) = \{ i \ \mid \ \phi(\mathbf{d})_i = 0 \ \& \ i \in [q] \}
\end{equation}
which collects the indices of all integer variables fixed at zero and
define function $\psi_i$ $(i \in [q])$ as
\begin{equation}
  \psi_i(\mathbf{d}) =
  \begin{cases}
    0 \mbox{ if } i \in S^0(\mathbf{d}) \\
    1 \mbox{ otherwise. }
  \end{cases}
\end{equation}
The learning task consists in designing a machine learning model
$\hat{\psi}_i$ that approximates $\psi_i$, i.e., $\hat{\psi}_i$ returns
the softmax probability of the outputs of $\psi_i$. The learning task
has at its disposal a distribution ${\cal D}$ of instances, which may
be obtained from historical data and/or forecasts.

\section{Supervised Learning}
\label{section:supervised}
To learn $S^0$, \propel{} uses a supervised learning approach and
generates $K$ instances $\bd_k \sim {\cal D}$ $(1 \leq k \leq
K)$ and their associated functions
$\psi_1(\bd_k),\ldots,\psi_q(\bd_k)$. The traditional way to train
$\hat{\psi}_i$ for a classification task is to use a cross-entropy
loss function:
\begin{equation*}
{\cal L}^c(\bd) = \psi_i(\bd) \log \hat{\psi}_i(\bd) + (1 - \psi_i(\bd)) \log (1 - \hat{\psi}_{i}(\bd)). 
\end{equation*}

\paragraph*{Instance Dependent Weights:}
Cross-entropy loss gauges the overall accuracy of the model, but it
does not directly incorporate the consequences of these predictions on
subsequent decision-making processes. Drawing inspiration from the
fraud detection literature, where each instance is weighted based on
transaction amount \citep{vanderschueren2022predict}, \propel{} scales
the weight of each training instance based on the optimal value of the
corresponding integer variable. This ensures that the learning process
emphasizes variables with larger values, which might be more critical
to the overall decision quality. Unlike class-dependent weights, which
simply shift the decision boundary to reduce expensive
misclassifications, instance-dependent weights vary with each instance
\citep{brefeld2003support}. \propel{} defines the weights of false
positives and false negatives as follows:
\begin{equation}
\begin{aligned}
& w_i^{\text{FP}} = 1 &\text{ } \\
\nonumber& w_i^{\text{FN}} = \exp\left(\frac{\psi_i(\bd)}{\sum_{j=1}^{q} \psi_j(\bd)}\right)& \label{weights}
\end{aligned}
\end{equation}
The loss function ${\cal L}^w(\bd)$ used in \propel{} for an instance
$\bd$ then becomes
\begin{equation*}
w_i^{\text{FN}} \psi_i(\bd) \log \hat{\psi}_i(\bd) + w_i^{\text{FP}} (1 - \psi_i(\bd)) \log (1 - \hat{\psi}_{i}(\bd)) 
\end{equation*}
and the overall loss function is defined as
\begin{equation}
  {\cal L} = \frac{1}{K} \sum_{k=1}^K {\cal L}^w(\bd_k).
\end{equation}

\paragraph*{Scoring with Reduced Costs:}

After training, $\hat{\psi}_i(\bd)$ gives the softmax probability that
variable $x_i$ is assigned to zero or a non-zero value in the optimal
solution $\psi(\bd)$. Denote by $\hat{\psi}_i^0(\bd) = \hat{\psi}_i(\bd)_0$ the
probability that $x_i$ be given the value 0 for an unseen instance $\bd$.
This probability can be used to approximate the set $S^0(\bd)$ as follows:
\begin{equation}
  \hat{S}^0(\bd) = \{ i \mid \hat{\psi}_i^0(\bd) \geq \tau \},
\end{equation}
where $\tau$ is a hyper-parameter threshold.

\propel{} refines this approximation by using the reduced costs in the
linear relaxation of the MIP problem; indeed, reduced cost provides
valuable information regarding whether a variable is likely to be
strictly positive.  Denote by $\underline{\phi}(\bd)$ the linear
relaxation of $\phi(\bd)$ and let $rc_i(\bd)$ be the reduced cost of
$x_i$ in $\underline{\phi}(\bd)$. \propel{} computes a normalized reduced
cost $r_i$ as follows:
\begin{equation}
r_i(\bd) = -\frac{1}{\pi} \arctan\left(\frac{rc_i(\bd)}{s}\right), \label{weighted_rc}
\end{equation}
where $s$ is a scaling factor. For instance, choosing $s$ as maximum
absolute value of all reduced costs ensures that $ -0.25 \le r_i\le
0.25$. \propel{} then approximates $S^0(\bd)$ as follows:
\begin{equation}
\hat{S}^0(\bd) = \{ i \mid  \hat{\psi}_i^0(\bd) + r_i(\bd) \geq \tau \}.
\end{equation}

\paragraph*{The Reduced MIP Model:} \propel{} uses the approximation $\hat{S}^0(\bd)$ to define a reduced MIP model. Let $\mathbf{d} = (\textbf{A}, \textbf{b}, \textbf{c})$ be a MIP instance. The reduced MIP is defined as follows:
\begin{equation}
\begin{aligned}
& \tilde{\phi}(\mathbf{d}) = & & \argmax_{\textbf{x}} &
  & \textbf{c}^T \textbf{x} \\ & & & \text{subject to} &
  & \textbf{Ax} \leq \textbf{b} \\ & & & & & x_i = 0 \;\;
  (i \in \hat{S}^0(\bd)) \\ & & & &
  & \textbf{x} \in \mathbb{Z}^q \times \mathbb{R}^{n-q} \label{Reduced-MIP}
\end{aligned}
\end{equation}
This paper assumes that the reduced MIP model has a feasible solution
(which is the case in the case study). This assumption is discussed in
detail in Section \ref{feasAssure}. 

\paragraph*{The Learning Architecture}

\propel{} uses a deep neural network architecture that was tuned by a
grid search. The details of the DNN architecture employed for the case
study are presented subsequently. Other ML models (e.g.,
SVM, Random Forest, etc.) have been evaluated, but have not
produced any benefit in either prediction or decision performance.

\section{Reinforcement Learning}
\label{section:DRL}
The supervised learning classification typically leads to reduced MIP
models whose solutions are close to optimality for the original
problem. However, occasionally, some instances exhibit optimality gaps
above the target value. To remedy this limitation, \propel{} adds a
Deep Reinforcement Learning (DRL) component to reduce the set of fixed
variables and obtain higher quality solutions.

\paragraph*{The Learning Goal}
\propel{} first partitions the fixed integer variables into subsets
$V_1,\ldots,V_m$; all or none of the variables in a subset $V_i$ will
be unfixed by the DRL component. Such partitions are natural for supply
chain planning applications, where variables often exhibit demand
patterns influenced by seasonal and market trends. For the industrial
case study presented later in the paper, \propel{} strategically
groups integer variables according to their temporal characteristics
and each subset $V_i$ corresponds to a different segment of the
planning horizon. This division leverages the temporal nature of the
variables, allowing the reintroduction of temporally related variables
into the model.

The goal of the DRL component is to find a subset
$J \subseteq \{1,\ldots,m\}$ that defines a reduced MIP:
\begin{equation}
\begin{aligned}
& \argmax_{\textbf{x}} & &  \textbf{c}^T \textbf{x} \\
&  \text{subject to} & & \textbf{Ax} \leq \textbf{b} \\
& & & x_i = 0 \;\; (i \in \hat{S}^0(\bd) \setminus \bigcup_{j \in J} V_j) \\
& & & \textbf{x} \in \mathbb{Z}^q \times \mathbb{R}^{n-q} 
\end{aligned}
\end{equation}
meeting the optimality target. The DRL component runs a number of
episodes, during which each step may ``unfix'' some subsets and
reinsert them into the MIP producing a reward, i.e., a better
objective value. After each episode, the DRL component trains a deep
learning network whose goal is to approximate an optimal policy that
determines which subsets to reinsert for unseen instances.

\paragraph*{The Model}
The DRL component is modeled as a Markov Decision Process (MDP). The
states of the MDP are of the form $\langle \mathbf{d},I,E \rangle$,
where $\mathbf{d}$ is the instance data, $I$ represents the sets that
are re-inserted, and $E$ represents the sets that are not selected for
re-insertion in the episode. Each state $s
= \langle \mathbf{d},I,E \rangle$ corresponds to the MIP $\tilde{\opt{}}(s)$
defined as
\begin{equation}
\begin{aligned}
& \max_{\textbf{x}} & &  \textbf{c}^T \textbf{x} \\
&  \text{subject to} & & \textbf{Ax} \leq \textbf{b} \\
& & & x_i = 0 \;\; (i \in \hat{S}^0(\bd) \setminus I) \\
& & & \textbf{x} \in \mathbb{Z}^q \times \mathbb{R}^{n-q} 
\end{aligned}
\end{equation}
The initial state $s_0$ is given by $\langle
\mathbf{d}, \emptyset, \emptyset \rangle$. A state $s$ is final if its optimal
solution $\app(s)$ is within the optimality tolerance.

The DRL component does not solve these MIPs to
optimality: rather it runs the optimization solver with a time limit
to determine whether reinserting some of the variable subsets leads to
an improved solution in reasonable time. 

Actions can be of two types, $\textsc{insert}(i)$ or
$\textsc{exclude}(i)$, which re-inserts or excludes a set $V_i$.
Given a state $s = \langle \mathbf{d},I,E \rangle$ and $i \in
[m] \setminus (I \cup E)$, the transitions are defined as follows:
\begin{equation*}
tr(\langle \mathbf{d},I,E \rangle,\textsc{insert}(i)) = \langle \mathbf{d},I \cup \{i\},E \rangle
\end{equation*}
\begin{equation*}
tr(\langle \mathbf{d},I,E \rangle,\textsc{exclude}(i)) = \langle \mathbf{d},I,E \cup \{i\} \rangle
\end{equation*}
The set of available actions in state $s$ is denoted by ${\cal A}(s)$.
The reward $r(s_t,s_{t+1})$ of a transition is defined by $\tilde{\cal
O}(s_{t+1})$, the objective value of the reduced MIP approximation
associated with $s_{t+1}$ . A state $s_t$ is final if $t > T$,
limiting the number of steps in an episode, or if its optimality gap
is within the target tolerance.

Solving the MDP consists in finding a policy $\pi$ from states to
actions that maximizes the value function
\[
V_{\pi}(s_t) = \sum_{t=0}^T \gamma^t r(s_t,s_{t+1)}=tr(s_t,\pi(s_t))
\]
where $s_{t+1}=tr(s_t,\pi(s_t))$. $V_{\pi}(s_t)$ can be rewritten as
\[
V_{\pi}(s_t) = r(s_t,s_{t+1}) + \gamma V_{\pi}(s_{t+1}).
\]
and the action-value function $Q_{\pi}(s_t,\mathbf{a})$ is defined as
\[
Q_{\pi}(s,a) = r(s,s') + \gamma V_{\pi}(s') \mbox{ where } s' = tr(s,a). 
\]
The optimal policy $\pi^*$ is given by
\[
\pi^* = \argmax_{\pi} V_{\pi}(s_0)
\]
and the optimal value function and action-value function are defined
as $V^* = V_{\pi^*}$ and $Q^* = Q_{\pi^*}$. By Bellman optimality condition,
the optimal action-value function $Q^*(s,a)$ can be rewritten as
\[
Q^*(s,a) = r(s,s'=tr(s,a)) + \gamma \max_{a' \in {\cal A}(s')} Q^*(s',a').
\]
The DRL component seeks to approximate $Q_{\pi^*}$ using a deep neural
network denoted by $\hat{Q}_\theta$ where $\theta$ are the network
parameters. The formula
\[
\hat{Q}_\theta(s,a) = r(s,s'=tr(s,a)) + \gamma \max_{a' \in {\cal A}(s')} \hat{Q}_\theta(s',a')
\]
provides a natural way to train the parameters $\theta$.


\paragraph*{Training the Action-Value Function Estimator}

Let $R = \{ \langle s_j, a_j, r_j, s_{j+1} \rangle \}_{j \in [|R|]}$
be a training set.  The training of the action-value function
estimator $\hat{Q}_\theta$ for $R$, denoted by $\textsc{Learn}(R,\hat{Q}_\theta)$,
consists in solving the following optimization problem
\begin{equation}
\label{eq:learning}
\argmin_{\theta} \frac{1}{|R|} \sum_{j=1}^{|R|} (y_j - \hat{Q}_\theta(s_j,a_j))^2
\end{equation}
where
\[
y_j = 
\begin{cases}
r_j  \mbox{ if } s_{j+1} \mbox{ is final;} \\
r_j + \gamma \max_{a \in {\cal A}(s_{j+1})} \hat{Q}_\theta(s_{j+1},a)  \mbox{ otherwise. }
\end{cases}
\]
The training algorithm is presented in
Algorithm \ref{alg:training}. It first initializes several
hyper-parameters, the learning parameters, the training set, and the
replay buffer.  It then runs multiple episodes, each with a specific
instance $\mathbf{d}$ from the training set.  During each episode,
the \propel{} training computes, and partitions,
$\hat{S}^0(\mathbf{d})$. It then runs $T_{\text{max}}$ steps, where
each step selects a random action with probability $\alpha$ or the
action with the best action-value approximation otherwise. \propel{}
then applies the action, computes its rewards, and updates the buffer.
At the end of the episode, \propel{} trains the action-value estimator
using \textsc{Learn}.

\begin{algorithm}[!t]
\caption{RL Training in \propel{}}
\label{alg:training}
\begin{algorithmic}[1]
\STATE \textbf{Initialize:}
\STATE \quad $T_{\text{max}}\gets$ maximum iterations per episode
\STATE \quad $\varepsilon \gets$ optimality gap threshold
\STATE \quad Initialize $\theta$ and $\gamma$
\STATE \quad ${\cal T} \gets$ training set with $N$ MIP instances 
\STATE \quad $R \gets \{\}$ \COMMENT{Replay Buffer}
\FOR{each episode in Episodes}
    \STATE select a MIP instance $\mathbf{d}$ from ${\cal T}$
    \STATE compute $\hat{S}^0(\mathbf{d})$ and partition it into $V_1,\ldots,V_m$
    \STATE $s_0 = \langle \mathbf{d},\emptyset,\emptyset \rangle$
    \FOR{$t = 0$ to $T_{\text{max}} - 1$}
        \STATE $\omega_{t} \gets $ optimality gap of $\hat{\cal O}(s_t)$
        \IF{$\omega_{t} \leq \varepsilon$}
            \STATE \textbf{break}
        \ENDIF
        \STATE {\bf with probability} $\alpha$, choose action $a_t$ randomly
        \STATE {\bf otherwise} $a_t = \max_{a \in {\cal A}(s_t)} \hat{Q}_{\theta}(s_t,a)$
        \STATE \textbf{apply action:} $s_{t+1} = tr(s_t,a_t)$
        \STATE \textbf{reward:} $r_{t} = r(s_t,s_{t+1})$
        \STATE \textbf{buffer update:} $R \gets R \cup \{ \langle s_t,a_t,r_t,s_{t+1} \rangle \}$
    \ENDFOR
    \STATE \textbf{learning:} $\theta \gets \textsc{Learn}(R,\hat{Q}_\theta)$
\ENDFOR
 \RETURN $\theta$
 \end{algorithmic}
\end{algorithm}

\paragraph*{Inference}

At inference time, \propel{} uses Algorithm \ref{alg:inference}. It
first predicts the set $\hat{S}^0(\mathbf{d})$ of fixed variables,
which is partitioned into $V_1,\ldots,V_m$. This provides the initial
state $s_0$. \propel{} then enters the DRL component and uses the
trained action-value function estimator $\hat{Q}_{\theta^*}$ to select
the next state in each iteration. The output is the state with the
best-found solution.

\begin{algorithm}[!t]
\caption{The Inference Algorithm of \propel{}.}
\label{alg:inference}
 \begin{algorithmic}[1]
     \STATE compute $\hat{S}^0(\mathbf{d})$ and partition it into $V_1,\ldots,V_m$
     \STATE $\mathbf{s}_0 = \langle d,\emptyset,\emptyset\rangle$ 
     \FOR{$t = 0$ to $T_{\text{max}}-1$}
         \STATE \textbf{action selection:} $\mathbf{a}_t = \max_{a_t \in {\cal A}(s_t)} \hat{Q}_{\theta^*}(s_t,a_t)$
         \STATE \textbf{apply action:} $s_{t+1} = tr(s_t,a_t)$
     \ENDFOR
     \RETURN $\argmax_{t \in [T_{\text{max}}-1]} \hat{\cal O}(s_t)$.
 \end{algorithmic}
\end{algorithm}

\paragraph*{Action Selection}

In practice, solving the optimization approximation for each
successive state may become computationally prohibitive, at training
and inference times. For this reason, \propel{} aggregates multiple
actions into a macro-action at each step. Instead of selecting the
best action at each step, macro-actions select all insertions seen as
beneficial based on their $Q$-value approximations, i.e.,
\begin{align*}
 {\cal M}(s) = \{ \textsc{insert}(i) \mid & \, i \in [m] \setminus (I \cup E) \; \& \\
&  \hat{Q}(s,\text{insert}(i)) \geq \hat{Q}(s,\text{exclude}(i)) \}
\end{align*}
The training is performed jointly for each action, i.e.,
Equation \eqref{eq:learning} becomes
\[
\argmin_{\theta}  \frac{1}{|R|} \sum_{j=1}^{|R|} \sum_{a \in {\cal A}(s_j)} (y_j - \hat{Q}_\theta(s_j,a))^2.
\]

\section{Case Study}
\label{section:SCP}
\label{section:formulation}

\propel{} was applied to several large-scale industrial supply chain
planning problems. The models are proprietary and, as a result, this
section presents a stylized version that captures most of the
realities in the field. The evaluations are based on the real
problems, which are particularly challenging.  Figure
\ref{fig:casestudy} specifies the parameters and decision variables of
the model. For each time period, the variables capture the demand met
for each product, the inventory held for each part, the quantity of
each part produced, and the unmet demand for each product.


Figure \ref{fig:model} presents the supply chain planning model.  The
objective function \eqref{7} minimizes the total cost which includes
the inventory holding costs (\(\alpha_{j}^t\)), the production costs
(\(\beta_{j}^t\)), and the unmet demand penalties
(\(\delta_{i}^t\)). Constraint \eqref{8} is the balance constraint; it
ensures that the inventory available at the end of each time period
\(t-1\) plus the production in time period \(t\) meets the demand in
\(t\) and the inventory requirements. Constraint \eqref{9} ensures
that the total quantity of demand \(D_{i}^t\) for each finished good
\(i\) at time \(t\) is either satisfied or accounted for as unmet
demand \(u_{i}^t\). Constraint \eqref{10} ensures that the total
production of all parts \(j \in m\) does not exceed the available
capacity \(\hat{P}_m^t\) for \(m \in \mathcal{C}_t\). These
constraints account for various capacity limits, such as time,
machinery, labor availability, etc.

\begin{figure}
\begin{tabular}{lp{4.5cm}}
\textbf{Sets \& Indices} & \\
\hline
\(i\) & Finished good index, \(i = 1, \ldots, M\) \\
\(j\) & Part index, \(j = 1, \ldots, N\) \\
\(t\) & Time index, \(t = 1, \ldots, T\) \\
\(\mathcal{S}_j\) & Set of demands (finished goods) that can be satisfied by supply \(j\) \\
\(\mathcal{C}^t\) & Set of production capacity resources at time period \(t\) \\
\(m\) & Capacity resource index \\ 
\(T_m\) & Part requiring capacity resource \(m\) 
\end{tabular}

\vspace{0.5cm}

\begin{tabular}{lp{4.5cm}}
\textbf{Input Parameters} & \\ 
\hline
\(D_{i}^t\) & demand for for \(i\) at time $t$ \\
\(\hat{P}_m^t\) & production capacity for $m$ at time $t$ \\
\(\alpha_{j}^t\) & inventory cost of part \(j\) at time \(t\) \\
\(\beta_{j}^t\) & production cost of part $j$ at time $t$ \\
\(\delta_{i}^t\) & penalty for unmet demand at time $t$ \\
\end{tabular}

\vspace{0.5cm}

\begin{tabular}{lp{4.5cm}}
\textbf{Decision Variables} & \\
\hline
\(x_{i}^t\) & demand for \(i\) met at time \(t\) \\
\(y_{j}^t\) & inventory of \(j\) after time period \(t\) \\
\(z_{j}^t\) & production of part \(j\) at time \(t\) \\
\(u_{i}^t\) & unmet demand for $i$ at time $t$
\end{tabular}\\
\caption{Description of the Inputs and Decision Variables.}
\label{fig:casestudy}
\end{figure}

The largest model considered in this study has approximately 1,143,576
rows, 6,140,652 nonzeros, and 2,151,770 columns containing 924,407
integer variables (0 binary). Given that both the production variables
($z_{j}^t$) and demand variables ($x_{i}^t$) are integers, the
inventory balance constraint \eqref{8} inherently causes the balance
variables $y_{j}^t$ to assume discrete values, even if they are
defined as continuous. Therefore, there is no need to explicitly
define them as integer variables. A similar rationale applies to the
unmet demand variables $u_{i}^t$ variables, which adopt discrete
values due to the integrality of the demand orders $D_{i}^t$.  For
further reading on advanced SCM mathematical modeling approaches,
refer to \citep{lee2016mathematical}.

\begin{figure}
\begin{align}
& \min \sum_{t=1}^{T} \sum_{j=1}^{N} \alpha_{j}^t y_{j}^t + \sum_{t=1}^{T} \sum_{j=1}^{N} \beta_{j}^t z_{j}^t + \sum_{t=1}^{T} \sum_{i=1}^{M} \delta_{i}^t u_{i}^t; \label{7}\\
\nonumber & \text{Subject to}\\
& y_{j}^{t-1} + z_{j}^t = \sum_{i\in \mathcal{S}_j} x_{i}^t + y_{j}^t \quad \forall j \in [N], \forall t \in [T] \label{8}\\
& \sum_{t=1}^{T} x_{i}^t + u_{i}^t = D_{i}^t \quad \forall i \in [M] \label{9}\\
& \sum_{j\in T_m} z_{j}^t \leq \hat{P}_m^t \quad \forall m \in \mathcal{C}^t, \forall t \in [T] \label{10}\\
& x_{i}^t, z_{j}^t \in\mathbb{Z} \quad \forall i \in [M], \forall j \in [N], \forall t \in [T] \label{11}\\
& y_{j}^t, u_{i}^t \ge 0 \quad \forall i \in [M], \forall j \in [N], \forall t \in [T]  \label{12}
\end{align}
\caption{The Supply Chain Planning Model.}
\label{fig:model}
\end{figure}


\paragraph*{Feasibility Assurance}
\label{feasAssure}

Addressing the feasibility of solutions in the context of
regression-based prediction approaches presents significant
challenges, particularly for equality constraints such as \eqref{8}
and \eqref{9}.  This difficulty arises from the inherent flexibility
within supply chain operations, where each demand $i$ can be satisfied
by various parts indexed by $j$. Consequently, each variable $x_i^t$
can be present in multiple balance constraints \eqref{8}. Therefore,
any adjustment to restore feasibility in one balance constraint can
inadvertently compromise the feasibility of many others, as these
constraints are interdependent within the network. This complexity is
significant because fixing a variable to a non-zero value incorrectly not only disrupts
individual constraints but may also cascade through the model,
affecting multiple balance equations. By predicting which variables
are fixed to zero rather than predicting their exact values, \propel{}
mitigates these propagation effects and is positioned to deliver
feasible, high-quality solutions.

The formulation of \propel{} in the previous section assumes that the
prediction phase ensures that the reduced MIP model
\eqref{Reduced-MIP} has a feasible solution. For the formulation
outlined in equations \eqref{7}-\eqref{12}, which models the practice
in the field, incorrect predictions do not render the problem
infeasible. Even if all integer variables are predicted to be fixed at
zero, implying no supply production and reliance solely on prior
inventory, the model maintains feasibility because unmet demands are
captured by the $u_{i}^t$ variables. These variables act as
slack variables in Lagrangian relaxation methods, which are often
used at the intersection of optimization and ML to manage
constraint violations \citep{fioretto2020predicting}.

If the model were formulated without the $u_{i}^t$ variables, thus
requiring all demands to be met without allowance for unmet demand,
the elimination of too many non-zero integer variables could lead to
infeasibility. In such scenarios, two mitigation strategies are
possible to apply \propel{}. The first strategy consists of adding
slack variables for each constraint and penalizing them by a
Lagrangian multiplier $\delta_{i}^t$ in the objective function.  The
other strategy is to generalize \propel{} so that the initial solution
can be infeasible. The DRL component then reinserts the variables to
restore both feasibility and near-optimality.

\paragraph*{Feature Extraction}

To learn function $\psi_i$, \propel{} does not use the entire dataset
$\mathbf{d}$. In most practical cases, SCP instances differ
primarily in the demand forecasts, i.e., the right-hand side of
Constraints \eqref{9}. However, the entire demand vector is not needed
for a specific variable in general. Limiting the number of input
features in a classifier can be advantageous for achieving superior
predictive performance and ensuring the model remains computationally
manageable \citep{zhang2000neural}. Moreover, the sizes of the feature
vectors can be further reduced by leveraging the inherent
characteristics of the problem.

For SCP optimization, \propel{} leverage the bipartite graph
representation of MIP models originally proposed
by \cite{gasse2019exact}. This graph representation encodes the
interactions between decision variables and constraints in the MIP
formulations. Let $G = (V, E)$ be a bipartite graph, where $V =
N \cup M = \{x_1, \ldots,
x_n\} \cup \{\kappa_{n+1}, \ldots, \kappa_{n+m}\}$ contains the set of
variable nodes ($N$) and constraint nodes ($M$). The edge set
$E \subseteq V \times V$ includes only those edges that connect nodes
of different types (variables and constraints), i.e., there is an edge
between variable $x_i$ and constraint $\kappa_j$ if $x_i$ appears in
$\kappa_j$.  Moreover, define $\rho_{ij} = 1$ if there exists a path
in $G$ between $x_i$ and $\kappa_j$ and $\rho_{ij} = 0$ otherwise, and
define $C(x_i) = \{\kappa_j \in V : \rho_{ij} = 1\}$ as the set of
constraints to consider for variable $x_i$. Indeed, whenever
$\kappa \in C(x)$, changes in the forecasted demand in $\kappa$ may
have an impact on $x$ and vice-versa. As a result, in a first
approximation, the features for $\psi_i$, denoted by $F(x_i)$ are
those forecasted demands in the constraints $C(x_i)$.

\begin{figure*}
\centering
\includegraphics[width=\linewidth]{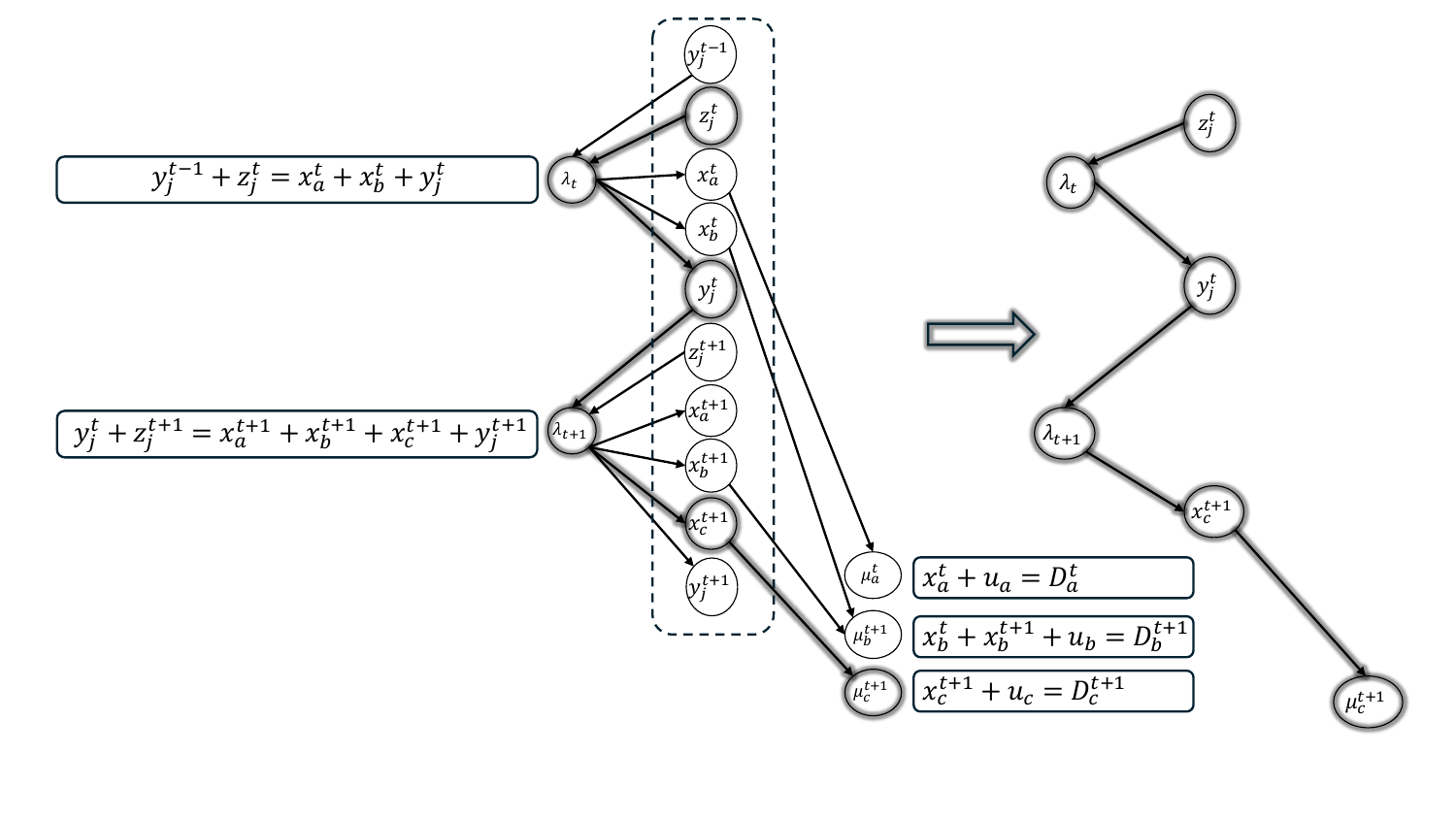}
\caption{\label{fig:graph}Transforming a MIP instance to a bipartite graph}
\end{figure*}

However, it is possible to exploit the structure of SCP applications
and, in particular, their temporal characteristics. In particular,
there should be no directed path from the future to the past. In this
context, an integer variable representing a supply on the day $t$
cannot be utilized to satisfy a demand due on a day $t^\prime <t$,
even if there exists an undirected path between them. This temporal
consideration significantly reduces the size of $F(x_i)$. Consider,
for instance, a 2-day planning horizon $(t, t+1)$, for part $j$ where
$\mathcal{S}_j = \{a,b,c\}$ and the due date for finished good $a$ is
time period $t$ and the due date for finished goods $b$ and $c$ are in
period $t+1$. Also, due to the single capacity constraint
(Constraint \eqref{10}), there is no production of part $c$ on day
$t$. Accordingly, the set of balance constraints \eqref{8} and demand
constraints \eqref{9} in the formulation can be expanded as follows:
\begin{align}
\nonumber& \text{Balance constraints:}\\
\nonumber& \quad\quad \lambda_t: \quad y_j^{t-1} + z_j^t - x_a^t - x_b^t - y_j^t = 0 \\
\nonumber& \quad\quad \lambda_{t+1}: \quad y_j^t + z_j^{t+1} - x_a^{t+1} - x_b^{t+1} - x_c^{t+1} - y_j^{t+1} = 0  \\
\nonumber& \text{Demand constraints:}\\
\nonumber& \quad\quad \mu_a^t: \quad x_a^t + u_a = D_a^t  \\
\nonumber& \quad\quad \mu_b^{t+1}: \quad x_b^t + x_b^{t+1} + u_b = D_b^{t+1} \\
\nonumber& \quad\quad \mu_c^{t+1}: \quad x_c^{t+1} + u_c = D_c^{t+1} 
\end{align}
Nodes and edges are graphically illustrated in
Figure \ref{fig:graph}. Variable nodes shown within the dashed box are
extracted from the variables.  Note that the graph is a bipartite
graph but for better visibility the balance constraints ($\lambda_{t}$
and $\lambda_{t+1}$) and demand constraints
($\mu_a^{t}, \mu_b^{t+1}, \text{and }\mu_c^{t+1}$) are depicted on the
left and right sides of the variable nodes, respectively. A directed
path from a variable ($z^t_j$) to constraint $\mu_c^{t+1}$ is
extracted from the graph as an example.  Similarly, after finding all
directed paths from integer variables to the demand constraints, the set
of features for each integer variable can be achieved, as listed
below:

\begin{align}
\nonumber& F(z_j^t) = \{D_a^t, D_b^{t+1}, D_c^{t+1}\}\\
\nonumber& F(x_a^t) = \{D_a^t\}  \\
\nonumber& F(z_j^{t+1}) = \{D_a^{t+1}, D_b^{t+1}, D_c^{t+1}\} \\
\nonumber& F(x_b^{t+1}) = \{D_b^{t+1}\}\\
\nonumber& F(x_a^{t+1}) = \emptyset \\
\nonumber& F(x_b^{t+1}) = \{D_b^{t+1}\} \\
\nonumber& F(x_c^{t+1}) = \{D_c^{t+1}\} 
\end{align}

\section{Computational Study}
\label{section:results}
This section provides a comprehensive computational analysis of
\propel{} for a large-scale SCM application, where decision making is
impacted by demand variability. \propel{}'s performance is evaluated
through a high-dimensional MIP case, showcasing its ability to
optimize complex SCP problems under a realistic industry scenario.

\subsection{Experimental Setting}
\paragraph*{Data Generation}

The model and datasets utilized in this study were sourced from
Kinaxis and encompass historical demand values. The base data is a set
of 20 snapshots taken throughout the year, each representing a
one-year planning horizon.  Each snapshot includes the demand for
product $i$ on a specific date $t$.

To generate training instances that are both representative and cover
realities in the field, historical demand values were perturbed to
simulate realistic fluctuations commonly observed in supply chain
data. In real-world settings, demand patterns exhibit both positive
correlations (due to trends like seasonality or complementary
products) and negative correlations (e.g., substitute goods). The
process to generate new instances consists of two steps.  The first
step selects a snapshot uniformly at random. Selecting a single
snapshot as a basis ensures that the generated training instances
capture not only the demand trends driven by seasonality but also
other influencing factors, such as real-world supply disruptions,
market dynamics, and evolving customer behaviors.  Once a snapshot is
selected, the second step perturbs the demand values for each product
separately. For each combination of part and time period, the data generation first applies a Gaussian noise $\epsilon \sim
\mathcal{N}(\mu,\sigma^2))$ with parameters $\mu$ and $\sigma$ and
then adds an additional uniform noise $\epsilon' \sim
U([-0.2,0.2])$. For all parts and time periods (about 300,000 data
points), the Gaussian distribution parameters, i.e., the mean and the
standard deviation, obey the following characteristics. The mean
$\mu$ across all parts/periods has an average value of $126,802.43$ and a
standard deviation of $427,862.92$.  The standard deviation has an
average of $68,439.84$ and a standard deviation of $194,076.55$. This captures, not only the magnitudes of the integer variables, but also their wide range,

\paragraph*{Computational Settings}

The training for supervised learning uses 500 MIP instances,
while the training of the DRL component uses 100 instances.  The
generation of the DRL training set takes place after the training of
the supervised learning phase. For training the DRL component, only
instances with a gap greater than the optimality tolerance are
selected, i.e., each instance $\mathbf{d}$ in the DRL training is such
that the reduced MIP $\tilde{\phi}(\mathbf{d})$ has an optimality gap
greater that the optimality tolerance. In the experiments, the
partition size for the DRL component is $8$.

The DNN models for supervised learning are multiple-layer perceptrons
with ReLU activations that were hyperparameter-tuned using a grid
search. The learning rates are taken from
$\{0.001,0.005\}$, the number of layers from $\{3, 4\}$, and the
hidden dimensions from $\{32, 64, 128\}$.  The input layer size is
determined by the number of features.  Each network is trained for 100
epochs with a batch size of 32. The model performances are evaluated
using F1 scores, along with confusion matrices. The Adam optimizer is
used to minimize the loss function with a learning rate
$\eta=0.005$. For every integer variable in Constraints
\eqref{7}-\eqref{12}, the best model is selected on the validation
set. The performance evaluations are on the test sets.

The Q-network is also a multiple-layer perceptron
with ReLU activations with two hidden layers of 128 neurons. 
The learning rate is 0.001, the discount factor is 0.99,
and the epsilon-greedy strategy uses $\epsilon = 0.1$

The training uses the ADAM optimizer with an initial learning rate of
0.005 over 100 epochs and employed a batch size of 32.  Models were
implemented using the PyTorch framework and trained on a Tesla V100
GPU hosted on Intel Xeon 2.7GHz machines. Predictions, optimization
solutions, and their corresponding objective values were obtained
using the Gurobi optimizer (Gurobi 10.0.1 (2023)) running on CPUs with
32 threads, with a termination criterion set by the Gurobi parameter
MIPGap = 1\%. The Gurobi parameter MIPFocus was set to one to
encourage finding feasible solutions more quickly.

%

\paragraph*{Metrics}
Given that the most effective predictive models do not necessarily
translate to optimal decisions in MIP
solving \citep{elmachtoub2020decision}, the focus of the paper is
exclusively on assessing the quality of decisions derived from
predictive models rather than the raw performance of the ML models.
Specifically, this assessment targets the architectural and conceptual
innovations of \propel{} as applied to the case study, exploring their
effectiveness through a comparative analysis with a conventional
optimization approach (denoted as ``OPT"). OPT employs the Gurobi
solver, leveraging its comprehensive capabilities, including presolve
routines, cutting planes, and heuristic solutions.

\subsection{Experimental Results}
The experimental evaluation is structured around two main analyses.
The first evaluation (Section
\ref{RCTest}) compares \prop{} and OPT when they are operating
under a runtime limit of 600 seconds.  The second evaluation
(Section \ref{reactExp}) extends this limit to 1000 seconds to assess
the full capabilities of \propel{} against OPT. The initial experiment
also investigates the impact of including reduced cost for improving
the predictive accuracy within this setting.\footnote{Note that the time
required to solve the LP relaxation, which is typically 10-12 seconds
for different instances of the case study, is included within the
total time limit. Therefore, if solving the LP relaxation takes 12
seconds, the subsequent model has only 588 seconds to run.} In
the \propel{} analysis, the solution of $\tilde{\phi}(\mathbf{d})$ after the
initial 600 seconds is used as a warm start in the DRL inference.  The
DRL inference performs up to four steps, each capped at 100
seconds. This process includes minor inference times, which are
negligible in the overall runtime.

\begin{table}
  \centering
  \caption{Comparative Analysis of Improvements and Reductions by \prop{}$^b$ and \prop{} against OPT.}
    \begin{tabular}{rlrr}
    \toprule
          & Method & \multicolumn{1}{c}{Max.} & \multicolumn{1}{c}{Avg.}  \\
    \midrule
    \multicolumn{1}{l}{Primal Integral} & PROP$^b$   & 20.20\% & 28.76\%  \\
          & PROP  & 45.76\% & 59.08\%  \\
    \midrule
    \multicolumn{1}{l}{Primal Gap} & PROP$^b$   & 56.45\% & 3.70\%  \\
          & PROP  & 56.64\% & 4.20\%  \\
    \midrule          
    \multicolumn{1}{l}{\# Integer Variables} & PROP$^b$   & 49.96\% & 32.06\%  \\
          & PROP  & 60.79\% & 48.79\%  \\
    \bottomrule
    \end{tabular}%
  \label{tab:summary1}%
\end{table}%



\subsection{Computational Performance of \prop{}}
\label{RCTest}

Table \ref{tab:summary1} summarizes the computational performance of
\prop{} against the baseline OPT model. It also performs an ablation
study and considers \prop{}$^b$, which is a version of \prop{} without
the reduced cost enhancement. Table \ref{tab:summary1} reports results
for the Primal Integral (PI), the Primal Gap (PG), and the reduction
in integer variables. For each metric, the maximum (Max.) and average
(Avg.) percentage improvements across for the
metric. The primal integral measures the quality of the solutions over
time, and higher reductions indicate better performance. Its
definition is given in Appendix \ref{section:metrics}.  The primal gap
assesses how close the solution is to optimality, and greater
reductions mean higher accuracy. Larger reductions in the number of
integer variables suggest more pruning of the search space.
Comprehensive details are shown in Table \ref{main_table}, which also
reports RunTimes (RT).

\vspace{1em}
\begin{table*}
  \centering
  \caption{Performance comparison of all three methods}
  \resizebox{\textwidth}{!}{%
    \renewcommand{\arraystretch}{1} 
    \begin{tabular}{llllllllllllllllll}
    \toprule
          & \multicolumn{5}{c}{OPT}               &       & \multicolumn{5}{c}{\prop{}$^b$}               &       & \multicolumn{5}{c}{PROP} \\
\cmidrule{2-6}\cmidrule{8-12}\cmidrule{14-18}    Instance & \multicolumn{1}{c}{ PI } &       & \multicolumn{1}{c}{PG} &       & \multicolumn{1}{c}{ RT } &       & \multicolumn{1}{c}{ PI } &       & \multicolumn{1}{c}{PG} &       & \multicolumn{1}{c}{ RT } &       & \multicolumn{1}{c}{ PI } &       & \multicolumn{1}{c}{PG} &       & \multicolumn{1}{c}{ RT } \\
    \midrule
    1     & \multicolumn{1}{r}{    130.90 } &       & \multicolumn{1}{r}{3.18\%} &       & \multicolumn{1}{r}{    600.14 } &       & \multicolumn{1}{r}{    100.21 } &       & \multicolumn{1}{r}{1.16\%} &       & \multicolumn{1}{r}{    600.14 } &       & \multicolumn{1}{r}{       75.99 } &       & \multicolumn{1}{r}{1.67\%} &       & \multicolumn{1}{r}{    600.13 } \\
    2     & \multicolumn{1}{r}{    134.51 } &       & \multicolumn{1}{r}{1.82\%} &       & \multicolumn{1}{r}{    600.21 } &       & \multicolumn{1}{r}{       90.87 } &       & \multicolumn{1}{r}{1.36\%} &       & \multicolumn{1}{r}{    600.36 } &       & \multicolumn{1}{r}{       66.24 } &       & \multicolumn{1}{r}{0.94\%} &       & \multicolumn{1}{r}{    241.96 } \\
    3     & \multicolumn{1}{r}{    125.51 } &       & \multicolumn{1}{r}{5.40\%} &       & \multicolumn{1}{r}{    600.11 } &       & \multicolumn{1}{r}{       60.42 } &       & \multicolumn{1}{r}{2.02\%} &       & \multicolumn{1}{r}{    600.13 } &       & \multicolumn{1}{r}{       44.21 } &       & \multicolumn{1}{r}{1.04\%} &       & \multicolumn{1}{r}{    241.55 } \\
    4     & \multicolumn{1}{r}{    117.23 } &       & \multicolumn{1}{r}{1.31\%} &       & \multicolumn{1}{r}{    600.15 } &       & \multicolumn{1}{r}{       71.05 } &       & \multicolumn{1}{r}{0.76\%} &       & \multicolumn{1}{r}{    206.78 } &       & \multicolumn{1}{r}{       36.06 } &       & \multicolumn{1}{r}{0.34\%} &       & \multicolumn{1}{r}{       36.15 } \\
    5     & \multicolumn{1}{r}{    122.69 } &       & \multicolumn{1}{r}{1.15\%} &       & \multicolumn{1}{r}{    600.12 } &       & \multicolumn{1}{r}{       41.64 } &       & \multicolumn{1}{r}{0.75\%} &       & \multicolumn{1}{r}{    123.72 } &       & \multicolumn{1}{r}{       33.18 } &       & \multicolumn{1}{r}{1.05\%} &       & \multicolumn{1}{r}{    106.85 } \\
    6     & \multicolumn{1}{r}{       55.32 } &       & \multicolumn{1}{r}{0.18\%} &       & \multicolumn{1}{r}{    198.78 } &       & \multicolumn{1}{r}{       44.80 } &       & \multicolumn{1}{r}{0.02\%} &       & \multicolumn{1}{r}{       72.07 } &       & \multicolumn{1}{r}{       25.71 } &       & \multicolumn{1}{r}{0.02\%} &       & \multicolumn{1}{r}{       25.80 } \\
    7     & \multicolumn{1}{r}{    103.47 } &       & \multicolumn{1}{r}{4.68\%} &       & \multicolumn{1}{r}{    600.11 } &       & \multicolumn{1}{r}{       87.09 } &       & \multicolumn{1}{r}{0.75\%} &       & \multicolumn{1}{r}{    312.91 } &       & \multicolumn{1}{r}{       62.23 } &       & \multicolumn{1}{r}{0.44\%} &       & \multicolumn{1}{r}{    188.47 } \\
    8     & \multicolumn{1}{r}{    139.31 } &       & \multicolumn{1}{r}{0.39\%} &       & \multicolumn{1}{r}{    311.40 } &       & \multicolumn{1}{r}{    120.10 } &       & \multicolumn{1}{r}{0.63\%} &       & \multicolumn{1}{r}{    230.51 } &       & \multicolumn{1}{r}{       33.26 } &       & \multicolumn{1}{r}{0.01\%} &       & \multicolumn{1}{r}{       33.36 } \\
    9     & \multicolumn{1}{r}{    145.53 } &       & \multicolumn{1}{r}{7.51\%} &       & \multicolumn{1}{r}{    600.10 } &       & \multicolumn{1}{r}{    110.10 } &       & \multicolumn{1}{r}{9.23\%} &       & \multicolumn{1}{r}{    600.12 } &       & \multicolumn{1}{r}{       48.17 } &       & \multicolumn{1}{r}{0.66\%} &       & \multicolumn{1}{r}{    208.69 } \\
    10    & \multicolumn{1}{r}{    143.99 } &       & \multicolumn{1}{r}{1.25\%} &       & \multicolumn{1}{r}{    600.11 } &       & \multicolumn{1}{r}{    101.50 } &       & \multicolumn{1}{r}{1.32\%} &       & \multicolumn{1}{r}{    600.11 } &       & \multicolumn{1}{r}{       39.28 } &       & \multicolumn{1}{r}{0.65\%} &       & \multicolumn{1}{r}{    139.87 } \\
    \midrule
    11    & \multicolumn{1}{r}{    146.06 } &       & \multicolumn{1}{r}{0.57\%} &       & \multicolumn{1}{r}{    316.83 } &       & \multicolumn{1}{r}{    118.13 } &       & \multicolumn{1}{r}{0.53\%} &       & \multicolumn{1}{r}{    229.18 } &       & \multicolumn{1}{r}{       32.77 } &       & \multicolumn{1}{r}{0.69\%} &       & \multicolumn{1}{r}{       32.87 } \\
    12    & \multicolumn{1}{r}{    113.77 } &       & \multicolumn{1}{r}{2.07\%} &       & \multicolumn{1}{r}{    600.10 } &       & \multicolumn{1}{r}{       82.73 } &       & \multicolumn{1}{r}{0.91\%} &       & \multicolumn{1}{r}{    343.56 } &       & \multicolumn{1}{r}{       73.34 } &       & \multicolumn{1}{r}{0.51\%} &       & \multicolumn{1}{r}{    192.99 } \\
    13    & \multicolumn{1}{r}{       89.16 } &       & \multicolumn{1}{r}{0.22\%} &       & \multicolumn{1}{r}{    239.39 } &       & \multicolumn{1}{r}{       78.40 } &       & \multicolumn{1}{r}{0.27\%} &       & \multicolumn{1}{r}{    174.79 } &       & \multicolumn{1}{r}{       32.02 } &       & \multicolumn{1}{r}{0.25\%} &       & \multicolumn{1}{r}{       32.11 } \\
    14    & \multicolumn{1}{r}{       84.62 } &       & \multicolumn{1}{r}{0.49\%} &       & \multicolumn{1}{r}{    248.81 } &       & \multicolumn{1}{r}{       64.40 } &       & \multicolumn{1}{r}{0.56\%} &       & \multicolumn{1}{r}{    172.28 } &       & \multicolumn{1}{r}{       31.37 } &       & \multicolumn{1}{r}{1.04\%} &       & \multicolumn{1}{r}{       31.47 } \\
    15    & \multicolumn{1}{r}{    119.54 } &       & \multicolumn{1}{r}{0.93\%} &       & \multicolumn{1}{r}{    476.93 } &       & \multicolumn{1}{r}{       79.52 } &       & \multicolumn{1}{r}{0.35\%} &       & \multicolumn{1}{r}{    246.73 } &       & \multicolumn{1}{r}{       35.06 } &       & \multicolumn{1}{r}{0.87\%} &       & \multicolumn{1}{r}{       35.16 } \\
    16    & \multicolumn{1}{r}{    195.90 } &       & \multicolumn{1}{r}{11.17\%} &       & \multicolumn{1}{r}{    600.11 } &       & \multicolumn{1}{r}{    156.32 } &       & \multicolumn{1}{r}{7.92\%} &       & \multicolumn{1}{r}{    600.12 } &       & \multicolumn{1}{r}{       58.79 } &       & \multicolumn{1}{r}{1.03\%} &       & \multicolumn{1}{r}{    567.63 } \\
    17    & \multicolumn{1}{r}{    162.56 } &       & \multicolumn{1}{r}{12.86\%} &       & \multicolumn{1}{r}{    600.11 } &       & \multicolumn{1}{r}{    125.27 } &       & \multicolumn{1}{r}{1.14\%} &       & \multicolumn{1}{r}{    600.12 } &       & \multicolumn{1}{r}{       84.13 } &       & \multicolumn{1}{r}{0.54\%} &       & \multicolumn{1}{r}{    148.05 } \\
    18    & \multicolumn{1}{r}{    145.89 } &       & \multicolumn{1}{r}{4.49\%} &       & \multicolumn{1}{r}{    600.12 } &       & \multicolumn{1}{r}{    122.41 } &       & \multicolumn{1}{r}{0.72\%} &       & \multicolumn{1}{r}{    431.64 } &       & \multicolumn{1}{r}{       63.59 } &       & \multicolumn{1}{r}{0.15\%} &       & \multicolumn{1}{r}{    120.16 } \\
    19    & \multicolumn{1}{r}{    158.27 } &       & \multicolumn{1}{r}{0.78\%} &       & \multicolumn{1}{r}{    486.64 } &       & \multicolumn{1}{r}{    122.49 } &       & \multicolumn{1}{r}{10.10\%} &       & \multicolumn{1}{r}{    600.13 } &       & \multicolumn{1}{r}{       45.43 } &       & \multicolumn{1}{r}{0.51\%} &       & \multicolumn{1}{r}{    210.51 } \\
    20    & \multicolumn{1}{r}{    139.04 } &       & \multicolumn{1}{r}{6.71\%} &       & \multicolumn{1}{r}{    600.12 } &       & \multicolumn{1}{r}{    120.68 } &       & \multicolumn{1}{r}{2.30\%} &       & \multicolumn{1}{r}{    600.10 } &       & \multicolumn{1}{r}{       78.28 } &       & \multicolumn{1}{r}{0.89\%} &       & \multicolumn{1}{r}{    233.84 } \\
    \midrule
    21    & \multicolumn{1}{r}{    118.73 } &       & \multicolumn{1}{r}{0.90\%} &       & \multicolumn{1}{r}{    325.45 } &       & \multicolumn{1}{r}{       85.26 } &       & \multicolumn{1}{r}{4.85\%} &       & \multicolumn{1}{r}{    600.10 } &       & \multicolumn{1}{r}{       34.51 } &       & \multicolumn{1}{r}{0.18\%} &       & \multicolumn{1}{r}{    117.28 } \\
    22    & \multicolumn{1}{r}{       88.86 } &       & \multicolumn{1}{r}{1.06\%} &       & \multicolumn{1}{r}{    347.70 } &       & \multicolumn{1}{r}{       66.21 } &       & \multicolumn{1}{r}{1.00\%} &       & \multicolumn{1}{r}{    308.44 } &       & \multicolumn{1}{r}{       33.79 } &       & \multicolumn{1}{r}{1.06\%} &       & \multicolumn{1}{r}{       94.87 } \\
    23    & \multicolumn{1}{r}{       88.08 } &       & \multicolumn{1}{r}{0.34\%} &       & \multicolumn{1}{r}{    242.64 } &       & \multicolumn{1}{r}{       52.99 } &       & \multicolumn{1}{r}{0.75\%} &       & \multicolumn{1}{r}{    225.97 } &       & \multicolumn{1}{r}{       30.44 } &       & \multicolumn{1}{r}{0.35\%} &       & \multicolumn{1}{r}{       30.54 } \\
    24    & \multicolumn{1}{r}{       85.84 } &       & \multicolumn{1}{r}{1.31\%} &       & \multicolumn{1}{r}{    600.12 } &       & \multicolumn{1}{r}{       60.50 } &       & \multicolumn{1}{r}{0.80\%} &       & \multicolumn{1}{r}{    171.87 } &       & \multicolumn{1}{r}{       33.60 } &       & \multicolumn{1}{r}{0.40\%} &       & \multicolumn{1}{r}{    110.96 } \\
    25    & \multicolumn{1}{r}{       63.19 } &       & \multicolumn{1}{r}{0.25\%} &       & \multicolumn{1}{r}{    153.45 } &       & \multicolumn{1}{r}{       40.67 } &       & \multicolumn{1}{r}{0.58\%} &       & \multicolumn{1}{r}{    105.26 } &       & \multicolumn{1}{r}{       31.87 } &       & \multicolumn{1}{r}{1.01\%} &       & \multicolumn{1}{r}{       34.19 } \\
    26    & \multicolumn{1}{r}{       53.65 } &       & \multicolumn{1}{r}{0.36\%} &       & \multicolumn{1}{r}{    146.20 } &       & \multicolumn{1}{r}{       37.20 } &       & \multicolumn{1}{r}{0.31\%} &       & \multicolumn{1}{r}{       99.65 } &       & \multicolumn{1}{r}{       26.30 } &       & \multicolumn{1}{r}{0.17\%} &       & \multicolumn{1}{r}{       60.78 } \\
    27    & \multicolumn{1}{r}{       65.68 } &       & \multicolumn{1}{r}{0.57\%} &       & \multicolumn{1}{r}{    125.33 } &       & \multicolumn{1}{r}{       31.44 } &       & \multicolumn{1}{r}{1.15\%} &       & \multicolumn{1}{r}{       31.54 } &       & \multicolumn{1}{r}{       24.84 } &       & \multicolumn{1}{r}{0.68\%} &       & \multicolumn{1}{r}{       28.61 } \\
    28    & \multicolumn{1}{r}{       57.45 } &       & \multicolumn{1}{r}{0.31\%} &       & \multicolumn{1}{r}{    112.47 } &       & \multicolumn{1}{r}{       32.94 } &       & \multicolumn{1}{r}{0.34\%} &       & \multicolumn{1}{r}{       80.61 } &       & \multicolumn{1}{r}{       29.02 } &       & \multicolumn{1}{r}{0.26\%} &       & \multicolumn{1}{r}{       77.83 } \\
    29    & \multicolumn{1}{r}{    131.89 } &       & \multicolumn{1}{r}{0.92\%} &       & \multicolumn{1}{r}{    122.64 } &       & \multicolumn{1}{r}{       37.17 } &       & \multicolumn{1}{r}{0.32\%} &       & \multicolumn{1}{r}{       70.21 } &       & \multicolumn{1}{r}{       23.83 } &       & \multicolumn{1}{r}{0.26\%} &       & \multicolumn{1}{r}{       23.92 } \\
    30    & \multicolumn{1}{r}{    140.77 } &       & \multicolumn{1}{r}{9.82\%} &       & \multicolumn{1}{r}{    600.12 } &       & \multicolumn{1}{r}{    116.31 } &       & \multicolumn{1}{r}{2.57\%} &       & \multicolumn{1}{r}{    600.13 } &       & \multicolumn{1}{r}{       97.68 } &       & \multicolumn{1}{r}{1.02\%} &       & \multicolumn{1}{r}{    290.12 } \\
    \midrule
    31    & \multicolumn{1}{r}{    160.08 } &       & \multicolumn{1}{r}{14.51\%} &       & \multicolumn{1}{r}{    600.09 } &       & \multicolumn{1}{r}{    149.26 } &       & \multicolumn{1}{r}{7.42\%} &       & \multicolumn{1}{r}{    600.29 } &       & \multicolumn{1}{r}{       98.08 } &       & \multicolumn{1}{r}{0.52\%} &       & \multicolumn{1}{r}{    289.63 } \\
    32    & \multicolumn{1}{r}{    186.14 } &       & \multicolumn{1}{r}{12.11\%} &       & \multicolumn{1}{r}{    600.14 } &       & \multicolumn{1}{r}{    132.87 } &       & \multicolumn{1}{r}{2.43\%} &       & \multicolumn{1}{r}{    600.15 } &       & \multicolumn{1}{r}{       86.97 } &       & \multicolumn{1}{r}{2.23\%} &       & \multicolumn{1}{r}{    600.13 } \\
    33    & \multicolumn{1}{r}{    171.39 } &       & \multicolumn{1}{r}{7.40\%} &       & \multicolumn{1}{r}{    600.13 } &       & \multicolumn{1}{r}{    105.24 } &       & \multicolumn{1}{r}{1.33\%} &       & \multicolumn{1}{r}{    600.11 } &       & \multicolumn{1}{r}{       37.48 } &       & \multicolumn{1}{r}{0.11\%} &       & \multicolumn{1}{r}{       32.78 } \\
    34    & \multicolumn{1}{r}{       62.82 } &       & \multicolumn{1}{r}{53.23\%} &       & \multicolumn{1}{r}{    600.30 } &       & \multicolumn{1}{r}{       76.59 } &       & \multicolumn{1}{r}{0.76\%} &       & \multicolumn{1}{r}{    211.08 } &       & \multicolumn{1}{r}{       29.93 } &       & \multicolumn{1}{r}{0.11\%} &       & \multicolumn{1}{r}{       30.02 } \\
    35    & \multicolumn{1}{r}{    132.64 } &       & \multicolumn{1}{r}{10.11\%} &       & \multicolumn{1}{r}{    600.10 } &       & \multicolumn{1}{r}{       75.42 } &       & \multicolumn{1}{r}{15.39\%} &       & \multicolumn{1}{r}{    600.11 } &       & \multicolumn{1}{r}{       36.33 } &       & \multicolumn{1}{r}{0.08\%} &       & \multicolumn{1}{r}{       36.43 } \\
    36    & \multicolumn{1}{r}{    144.64 } &       & \multicolumn{1}{r}{9.98\%} &       & \multicolumn{1}{r}{    600.12 } &       & \multicolumn{1}{r}{    100.04 } &       & \multicolumn{1}{r}{0.94\%} &       & \multicolumn{1}{r}{    287.94 } &       & \multicolumn{1}{r}{       98.29 } &       & \multicolumn{1}{r}{2.34\%} &       & \multicolumn{1}{r}{    600.12 } \\
    37    & \multicolumn{1}{r}{    114.53 } &       & \multicolumn{1}{r}{0.76\%} &       & \multicolumn{1}{r}{    290.94 } &       & \multicolumn{1}{r}{    123.71 } &       & \multicolumn{1}{r}{3.85\%} &       & \multicolumn{1}{r}{    600.12 } &       & \multicolumn{1}{r}{       63.43 } &       & \multicolumn{1}{r}{0.68\%} &       & \multicolumn{1}{r}{    145.02 } \\
    38    & \multicolumn{1}{r}{    144.51 } &       & \multicolumn{1}{r}{9.96\%} &       & \multicolumn{1}{r}{    600.09 } &       & \multicolumn{1}{r}{    113.62 } &       & \multicolumn{1}{r}{0.79\%} &       & \multicolumn{1}{r}{    478.22 } &       & \multicolumn{1}{r}{       99.70 } &       & \multicolumn{1}{r}{0.88\%} &       & \multicolumn{1}{r}{    259.48 } \\
    39    & \multicolumn{1}{r}{       62.28 } &       & \multicolumn{1}{r}{57.32\%} &       & \multicolumn{1}{r}{    600.30 } &       & \multicolumn{1}{r}{       82.70 } &       & \multicolumn{1}{r}{0.87\%} &       & \multicolumn{1}{r}{    126.19 } &       & \multicolumn{1}{r}{       39.00 } &       & \multicolumn{1}{r}{0.68\%} &       & \multicolumn{1}{r}{       98.06 } \\
    40    & \multicolumn{1}{r}{    147.96 } &       & \multicolumn{1}{r}{13.62\%} &       & \multicolumn{1}{r}{    600.13 } &       & \multicolumn{1}{r}{    124.63 } &       & \multicolumn{1}{r}{2.04\%} &       & \multicolumn{1}{r}{    600.14 } &       & \multicolumn{1}{r}{       51.07 } &       & \multicolumn{1}{r}{0.78\%} &       & \multicolumn{1}{r}{    161.96 } \\
    \midrule
    41    & \multicolumn{1}{r}{    115.50 } &       & \multicolumn{1}{r}{0.60\%} &       & \multicolumn{1}{r}{    283.82 } &       & \multicolumn{1}{r}{       98.69 } &       & \multicolumn{1}{r}{0.96\%} &       & \multicolumn{1}{r}{    189.99 } &       & \multicolumn{1}{r}{       33.85 } &       & \multicolumn{1}{r}{0.49\%} &       & \multicolumn{1}{r}{    152.54 } \\
    42    & \multicolumn{1}{r}{    194.43 } &       & \multicolumn{1}{r}{15.95\%} &       & \multicolumn{1}{r}{    600.17 } &       & \multicolumn{1}{r}{    137.52 } &       & \multicolumn{1}{r}{2.39\%} &       & \multicolumn{1}{r}{    600.12 } &       & \multicolumn{1}{r}{       35.14 } &       & \multicolumn{1}{r}{0.93\%} &       & \multicolumn{1}{r}{       35.24 } \\
    43    & \multicolumn{1}{r}{    195.81 } &       & \multicolumn{1}{r}{3.95\%} &       & \multicolumn{1}{r}{    600.15 } &       & \multicolumn{1}{r}{    132.04 } &       & \multicolumn{1}{r}{12.20\%} &       & \multicolumn{1}{r}{    600.12 } &       & \multicolumn{1}{r}{       64.10 } &       & \multicolumn{1}{r}{0.75\%} &       & \multicolumn{1}{r}{    247.04 } \\
    44    & \multicolumn{1}{r}{    147.20 } &       & \multicolumn{1}{r}{2.47\%} &       & \multicolumn{1}{r}{    600.12 } &       & \multicolumn{1}{r}{    108.06 } &       & \multicolumn{1}{r}{4.94\%} &       & \multicolumn{1}{r}{    600.11 } &       & \multicolumn{1}{r}{       81.60 } &       & \multicolumn{1}{r}{1.14\%} &       & \multicolumn{1}{r}{    600.13 } \\
    45    & \multicolumn{1}{r}{    132.28 } &       & \multicolumn{1}{r}{1.46\%} &       & \multicolumn{1}{r}{    600.09 } &       & \multicolumn{1}{r}{    122.97 } &       & \multicolumn{1}{r}{0.71\%} &       & \multicolumn{1}{r}{    522.14 } &       & \multicolumn{1}{r}{       53.80 } &       & \multicolumn{1}{r}{2.35\%} &       & \multicolumn{1}{r}{    600.11 } \\
    46    & \multicolumn{1}{r}{    139.78 } &       & \multicolumn{1}{r}{1.23\%} &       & \multicolumn{1}{r}{    600.12 } &       & \multicolumn{1}{r}{    118.68 } &       & \multicolumn{1}{r}{3.94\%} &       & \multicolumn{1}{r}{    600.10 } &       & \multicolumn{1}{r}{       87.25 } &       & \multicolumn{1}{r}{0.91\%} &       & \multicolumn{1}{r}{    422.13 } \\
    47    & \multicolumn{1}{r}{    137.51 } &       & \multicolumn{1}{r}{2.71\%} &       & \multicolumn{1}{r}{    600.21 } &       & \multicolumn{1}{r}{       97.99 } &       & \multicolumn{1}{r}{1.52\%} &       & \multicolumn{1}{r}{    600.16 } &       & \multicolumn{1}{r}{       33.73 } &       & \multicolumn{1}{r}{1.07\%} &       & \multicolumn{1}{r}{       33.81 } \\
    48    & \multicolumn{1}{r}{    139.68 } &       & \multicolumn{1}{r}{14.17\%} &       & \multicolumn{1}{r}{    600.13 } &       & \multicolumn{1}{r}{       96.43 } &       & \multicolumn{1}{r}{0.10\%} &       & \multicolumn{1}{r}{    202.47 } &       & \multicolumn{1}{r}{       67.73 } &       & \multicolumn{1}{r}{1.01\%} &       & \multicolumn{1}{r}{    145.10 } \\
    49    & \multicolumn{1}{r}{    175.73 } &       & \multicolumn{1}{r}{17.49\%} &       & \multicolumn{1}{r}{    600.12 } &       & \multicolumn{1}{r}{    147.14 } &       & \multicolumn{1}{r}{8.06\%} &       & \multicolumn{1}{r}{    600.15 } &       & \multicolumn{1}{r}{       71.12 } &       & \multicolumn{1}{r}{0.94\%} &       & \multicolumn{1}{r}{    364.34 } \\
    50    & \multicolumn{1}{r}{    133.81 } &       & \multicolumn{1}{r}{0.21\%} &       & \multicolumn{1}{r}{    213.57 } &       & \multicolumn{1}{r}{       93.69 } &       & \multicolumn{1}{r}{0.67\%} &       & \multicolumn{1}{r}{    196.00 } &       & \multicolumn{1}{r}{       67.54 } &       & \multicolumn{1}{r}{0.10\%} &       & \multicolumn{1}{r}{    116.70 } \\
    \midrule
    51    & \multicolumn{1}{r}{       63.07 } &       & \multicolumn{1}{r}{1.65\%} &       & \multicolumn{1}{r}{    600.13 } &       & \multicolumn{1}{r}{       36.42 } &       & \multicolumn{1}{r}{0.22\%} &       & \multicolumn{1}{r}{       36.52 } &       & \multicolumn{1}{r}{       30.37 } &       & \multicolumn{1}{r}{0.67\%} &       & \multicolumn{1}{r}{       79.53 } \\
    52    & \multicolumn{1}{r}{    191.45 } &       & \multicolumn{1}{r}{12.11\%} &       & \multicolumn{1}{r}{    600.11 } &       & \multicolumn{1}{r}{       63.39 } &       & \multicolumn{1}{r}{1.12\%} &       & \multicolumn{1}{r}{    600.13 } &       & \multicolumn{1}{r}{       61.84 } &       & \multicolumn{1}{r}{2.10\%} &       & \multicolumn{1}{r}{    600.13 } \\
    53    & \multicolumn{1}{r}{    172.61 } &       & \multicolumn{1}{r}{1.87\%} &       & \multicolumn{1}{r}{    600.13 } &       & \multicolumn{1}{r}{       52.07 } &       & \multicolumn{1}{r}{3.76\%} &       & \multicolumn{1}{r}{    600.13 } &       & \multicolumn{1}{r}{       47.64 } &       & \multicolumn{1}{r}{0.08\%} &       & \multicolumn{1}{r}{       47.75 } \\
    54    & \multicolumn{1}{r}{    139.66 } &       & \multicolumn{1}{r}{12.65\%} &       & \multicolumn{1}{r}{    600.11 } &       & \multicolumn{1}{r}{       88.72 } &       & \multicolumn{1}{r}{0.70\%} &       & \multicolumn{1}{r}{    215.12 } &       & \multicolumn{1}{r}{       34.59 } &       & \multicolumn{1}{r}{0.11\%} &       & \multicolumn{1}{r}{       34.68 } \\
    55    & \multicolumn{1}{r}{    169.93 } &       & \multicolumn{1}{r}{5.80\%} &       & \multicolumn{1}{r}{    600.11 } &       & \multicolumn{1}{r}{       38.59 } &       & \multicolumn{1}{r}{0.64\%} &       & \multicolumn{1}{r}{    116.11 } &       & \multicolumn{1}{r}{       39.41 } &       & \multicolumn{1}{r}{0.42\%} &       & \multicolumn{1}{r}{    215.11 } \\
    56    & \multicolumn{1}{r}{    182.91 } &       & \multicolumn{1}{r}{3.02\%} &       & \multicolumn{1}{r}{    600.11 } &       & \multicolumn{1}{r}{    151.26 } &       & \multicolumn{1}{r}{1.52\%} &       & \multicolumn{1}{r}{    600.11 } &       & \multicolumn{1}{r}{       94.24 } &       & \multicolumn{1}{r}{3.30\%} &       & \multicolumn{1}{r}{    600.11 } \\
    57    & \multicolumn{1}{r}{    142.04 } &       & \multicolumn{1}{r}{13.43\%} &       & \multicolumn{1}{r}{    600.14 } &       & \multicolumn{1}{r}{    138.77 } &       & \multicolumn{1}{r}{7.09\%} &       & \multicolumn{1}{r}{    600.11 } &       & \multicolumn{1}{r}{       79.54 } &       & \multicolumn{1}{r}{0.40\%} &       & \multicolumn{1}{r}{    249.81 } \\
    58    & \multicolumn{1}{r}{    135.65 } &       & \multicolumn{1}{r}{1.53\%} &       & \multicolumn{1}{r}{    600.15 } &       & \multicolumn{1}{r}{    130.27 } &       & \multicolumn{1}{r}{1.95\%} &       & \multicolumn{1}{r}{    600.11 } &       & \multicolumn{1}{r}{    106.26 } &       & \multicolumn{1}{r}{0.24\%} &       & \multicolumn{1}{r}{    309.47 } \\
    59    & \multicolumn{1}{r}{    146.44 } &       & \multicolumn{1}{r}{4.52\%} &       & \multicolumn{1}{r}{    600.11 } &       & \multicolumn{1}{r}{       67.81 } &       & \multicolumn{1}{r}{1.23\%} &       & \multicolumn{1}{r}{    600.31 } &       & \multicolumn{1}{r}{       55.01 } &       & \multicolumn{1}{r}{0.84\%} &       & \multicolumn{1}{r}{    296.38 } \\
    60    & \multicolumn{1}{r}{    132.72 } &       & \multicolumn{1}{r}{1.01\%} &       & \multicolumn{1}{r}{    298.51 } &       & \multicolumn{1}{r}{       80.81 } &       & \multicolumn{1}{r}{0.97\%} &       & \multicolumn{1}{r}{    193.16 } &       & \multicolumn{1}{r}{       34.87 } &       & \multicolumn{1}{r}{0.78\%} &       & \multicolumn{1}{r}{    161.46 } \\
    \midrule
    \multicolumn{18}{l}{\% PI: Primal Integral, PG: Primal Gap, RT: Run Time} \\
    \end{tabular}%
    }
  \label{main_table}%
\end{table*}%

\begin{figure}
\centering
\includegraphics[width=\linewidth]{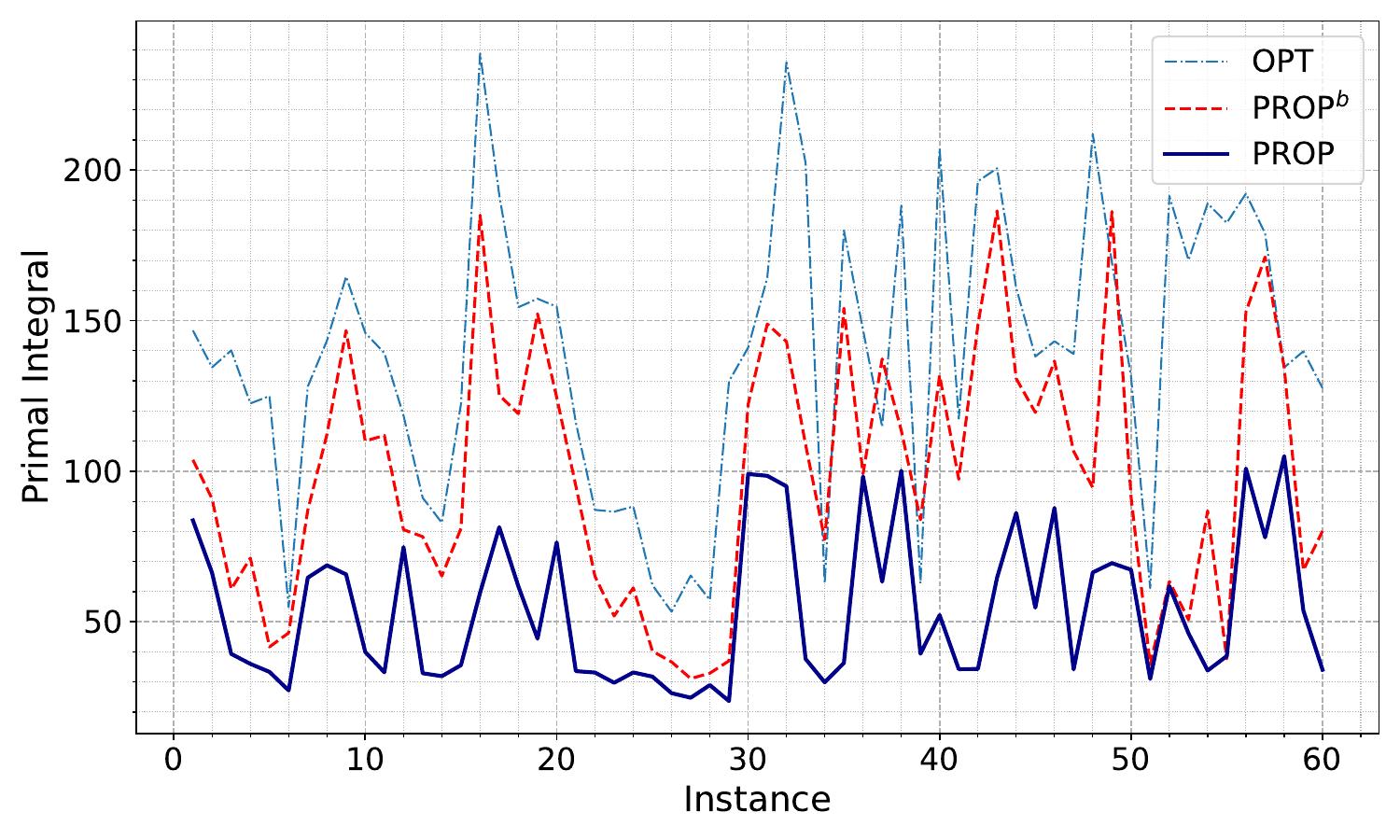}
\caption{\label{PI60} Comparison of Primal Integral Values Across All Instances.}
\end{figure}

\paragraph*{Primal Integral} The PI results first show that \prop{} consistently
achieves higher improvements across all metrics compared to
\prop{}$^b$, highlighting its superior efficiency and effectiveness.
{\em The inclusion of reduced costs in identifying which variables to
fix is a significant contribution.}  Second, both methods achieve
significant improvement over OPT. On average, \prop{}$^b$ achieves a
reduction of 28.76\% in the primal integral, while \prop{} reaches an
impressive 59.08\% reduction. Figure \ref{PI60} highlights these
benefits clearly: while \prop{}$^b$outperforms the OPT method in 57
out of 60 instances, \prop{} consistently delivers the best primal
integral across all instances.  Both \prop{}$^b$ and \prop{} exhibit
markedly better performance than the baseline, showcasing their
effectiveness. {\em \prop{} reduces the average primal integral over
the 60 instances by a factor of 5.72}, which is derived from the
reduction in magnitude of the primal integral values reported in the
table.  This substantial improvement underscores the robustness and
efficiency of \prop{} in improving the primal integral. Since \prop{}
significantly outperforms \prop{}$^b$, the comments in the following
focus on \prop{}.

\begin{figure}
\centering
\includegraphics[width=\linewidth]{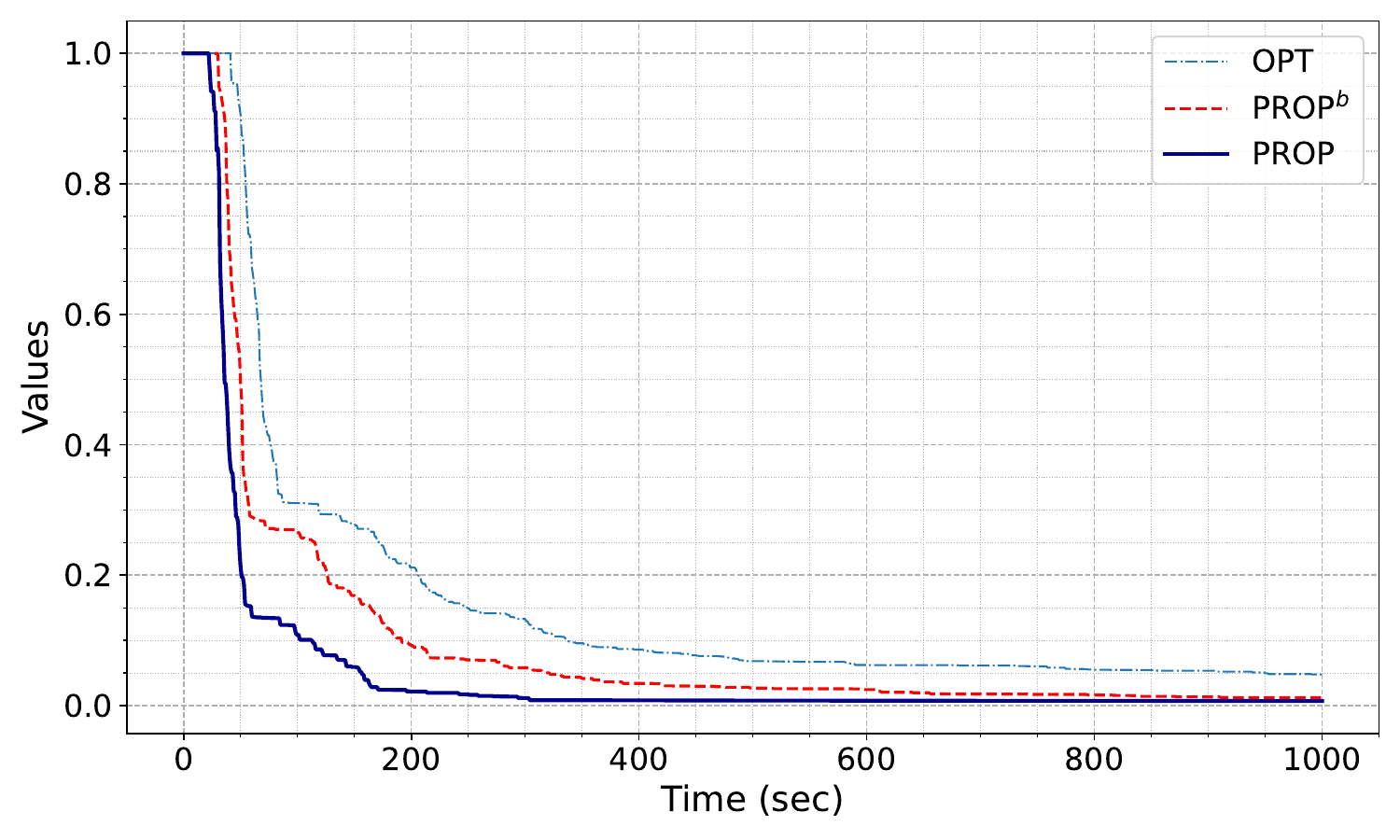}
\caption{\label{fig:Primal}Temporal Evolution of the Average Primal Gaps across 60 Test Instances.}
\end{figure}

\begin{figure}
\centering
\includegraphics[width=\linewidth]{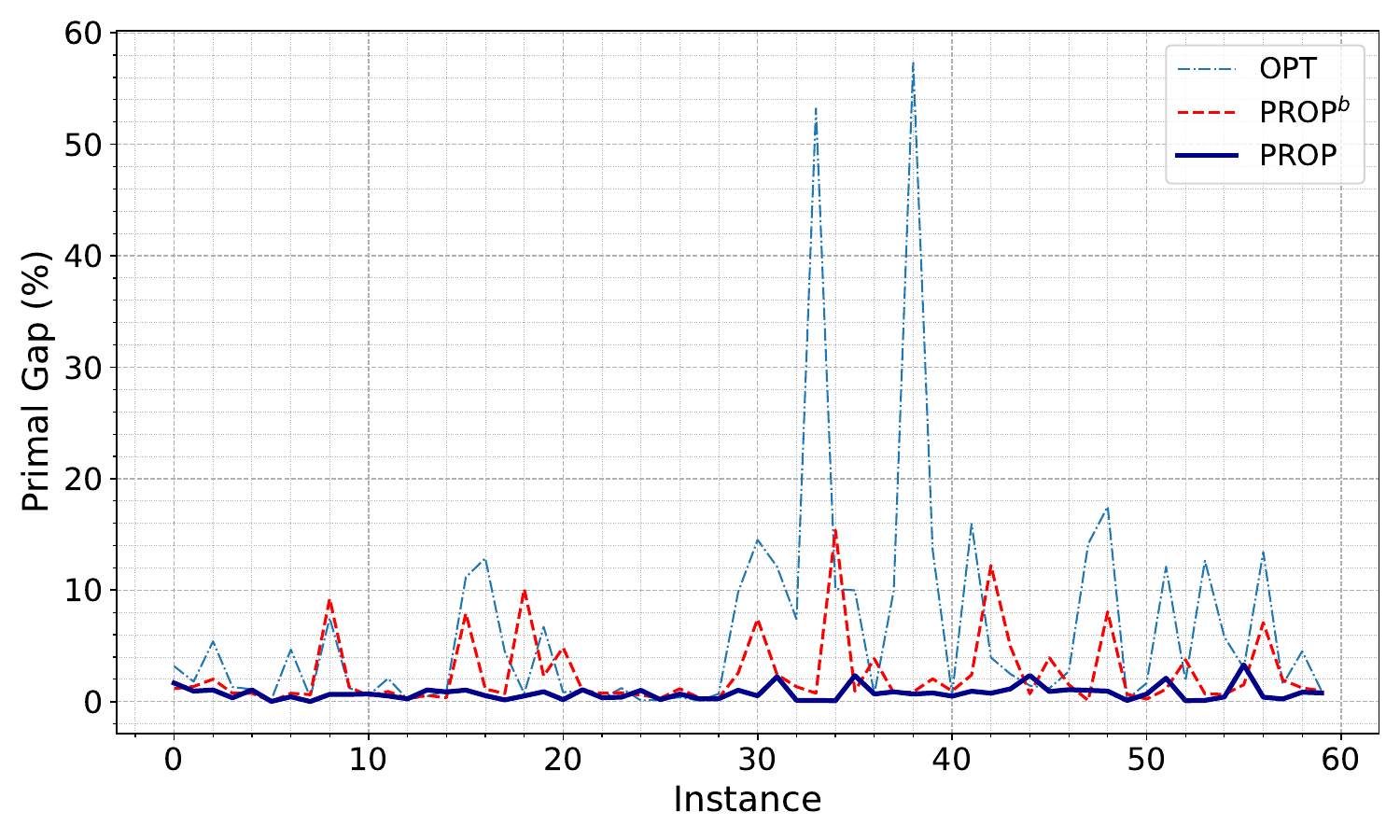}
\caption{\label{PGap}Primal Gap Values at the Time Limit $t_{max}$.}
\end{figure}

\paragraph*{Primal Gap}
Figure \ref{fig:Primal} provides a temporal view of the primal gap
trends throughout the operational timeline of 600 seconds across the
three methods. The values at each time $t \in [0, 600]$ are averaged
across 60 instances. \prop{} exhibit a much faster decay in the primal
gap than OPT and \prop{}$^b$. Table \ref{main_table} provides detailed
statistics: it shows that {\em \prop{} exhibit improvements
in the primal gap of 88.26\%. \prop{} has primal gaps
that are up to 15 times smaller than those of OPT.} Table
\ref{tab:summary1} and Figure \ref{PGap} detail these results.
In particular, Figure \ref{PGap} shows the dramatic improvements in
 primal gap of \prop{} over OPT, and highlights the consistency and
 robustness of \prop{}.  These improvements are largely attributed to
 the reduction in the number of integer
 variables. Table \ref{tab:summary1} shows that {\em \prop{} reduces
 the number of integer variables by a 48.79\% in average}, compared to
 the 32.06\% reduction by \prop{}$^b$. The reduction in integer
 variables and the corresponding decrease in computational complexity
 underscore the effectiveness of \prop{}.

\begin{table}
  \centering
  \caption{Time (sec) to the first incumbents in each method}
    \begin{tabular}{lrrr}
    \toprule
          & \multicolumn{1}{l}{OPT} & \multicolumn{1}{l}{PROP$^b$} & \multicolumn{1}{l}{PROP} \\
    \midrule
    Max.  & 118   & 73    & 83 \\
    Avg.  & 69    & 43    & 40 \\
    Min   & 37    & 25    & 23 \\
    \bottomrule
    \end{tabular}%
  \label{tab:inc}%
\end{table}%

\paragraph*{Solution Quality}

\begin{figure}
\centering
\includegraphics[width=\linewidth]{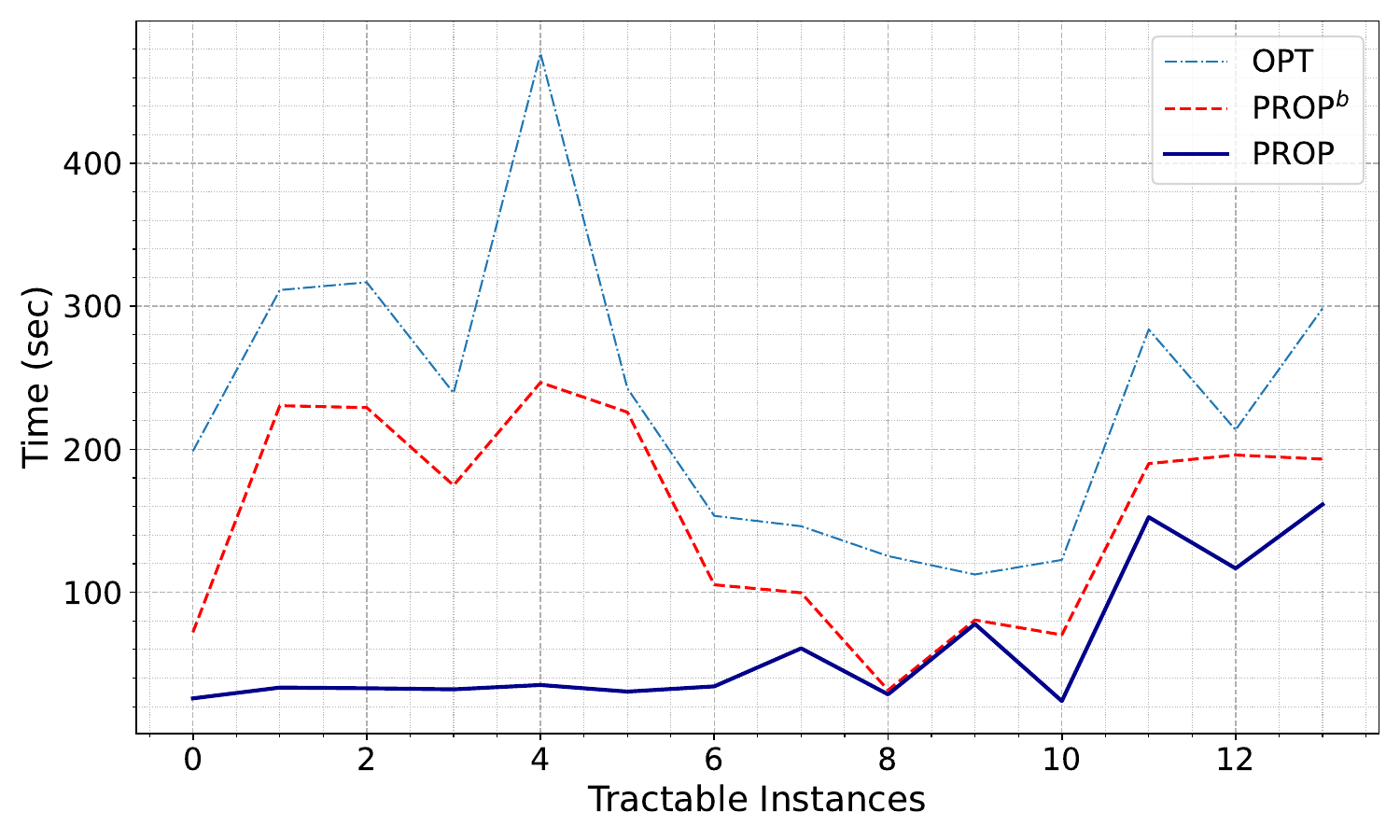}
\caption{\label{runtime_fig}Comparing Runtime Values in Tractable Instances Within the Time Limit.}
\end{figure}

The maximum, minimum, and average time values for the first incumbents
in each method are presented in Table \ref{tab:inc}. \prop{} improves
the average time to the first incumbents by 37.83\% over
OPT. Table \ref{main_table} shows that OPT failed to find a solution
within the optimality tolerance in 41 out of 60 test instances.  This
corresponds to a failure rate of 71.67\%. The failure rate is reduced
to 26.67\% (15 instances) for \prop{}.  Figure \ref{runtime_fig} and
Table \ref{tab:statistics_summary} present the solution times for the
14 instances that are solved within the time limit by all three
methods.  They indicate that {\em \prop{} significantly
outperform OPT. Specifically,
\prop{} achieves a 73.91\% reduction in mean solution time and a
substantial 85.09\% reduction in median solution time}. Additionally,
\prop{} achieves a greater consistency in solution times, as indicated
by the lower standard deviation, and significant reductions in both
minimum and maximum solution times.


\begin{table}
  \centering
  \caption{Comparative Analysis of Solution Times  for Instances Terminated before the Time Limit (600 sec)}
    \begin{tabular}{lrrr}
    \toprule
          & OPT & PROP$^b$ & PROP \\
    \midrule
    Mean    & 231.57 & 153.26 & 60.42 \\
    Std     & 101.29 & 73.25 & 48.21 \\
    Min     & 112.47 & 31.54 & 23.92 \\
    25\%    & 148.02 & 85.37 & 30.93 \\
    50\%    & 226.48 & 182.39 & 33.77 \\
    75\%    & 294.83 & 218.48 & 73.57 \\
    Max     & 476.93 & 246.73 & 161.46 \\
    \bottomrule
    \end{tabular}%
  \label{tab:statistics_summary}%
\end{table}%

In summary, the runtimes of \prop{} range from 1.44 to 13.57 times
faster than the other methods, with an average improvement factor of
5.64.  \prop{} achieves better primal gaps and terminates faster.

\subsection{The Benefits of \propel{} and Deep RL}
\label{reactExp}

This section focuses on the 15 instances where \prop{} fails to
achieve a primal gap below 1\% within the prescribed 600-second limit.
The experiments consider longer runtimes of 1,000 seconds for OPT
and \propel{}. \propel{} allocates 600 seconds to \prop{} and the rest
to the DRL component. Each iteration of the DRL component is given 100
seconds, including the minor inference times.

\begin{figure}
\centering
\includegraphics[width=\linewidth]{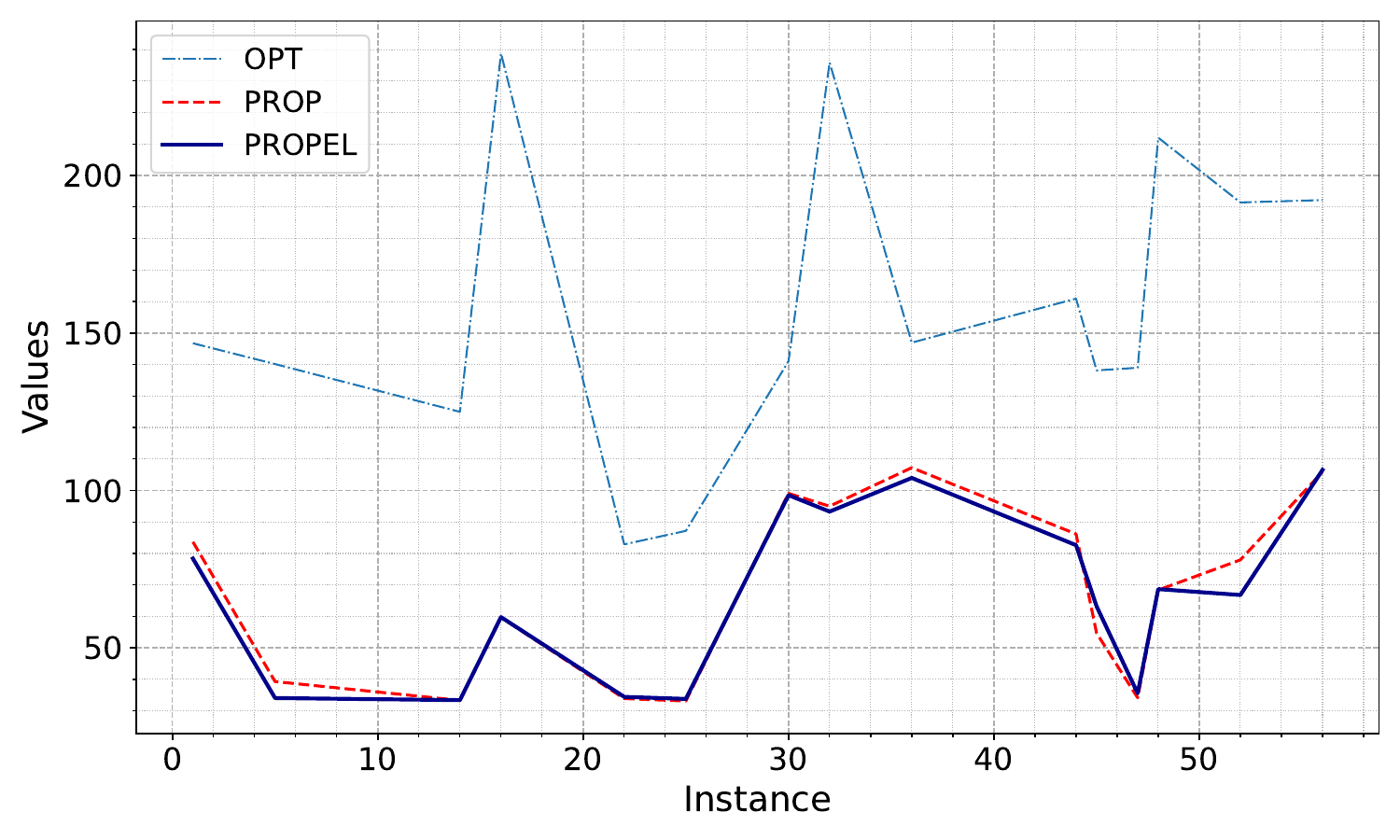}
\caption{\label{RL_AUC}Primal Integral Values After 1000 Seconds.}
\end{figure}

\paragraph*{Primal Integral}
Figure \ref{RL_AUC} visualizes the primal integral values
of \propel{}.  Additionally, a detailed comparison between primal
integral percentage reduction among the methods is provided in Table
\ref{tab:comparison_metrics}. The table shows 57.46\% and 58.24\%
reductions in primal gaps for \prop{} and \propel{}, respectively,
compared to the OPT. The results indicate that the RL component
enhances efficiency compared to merely extending the runtime of the
OPT and \prop{} methods. Detailed primal integral values for each
method are provided in Table \ref{tab:RL_AUC}.

\begin{table*}
  \centering
  \caption{Primal Integral Comparative Analysis within a 1000-second Time Limit.}
    \begin{tabular}{lrrrrr}
    \toprule
          & OPT & \prop{} & \propel{} & PROP\_Reduction (\%) & \propel{}\_Reduction (\%) \\
    \midrule
    Mean    & 158.55 & 67.44 & 66.21 & 57.46 & 58.24 \\
    Median  & 146.74 & 68.38 & 66.81 & 53.40 & 54.47 \\
    Std Dev & 47.18  & 28.20 & 27.26 & 40.23 & 42.21 \\
    Min     & 82.89  & 33.14 & 33.43 & 60.03 & 59.67 \\
    Max     & 238.78 & 107.23 & 106.52 & 55.09 & 55.39 \\
    \bottomrule
    \end{tabular}%
  \label{tab:comparison_metrics}%
\end{table*}%

\begin{figure}
\centering
\includegraphics[width=\linewidth]{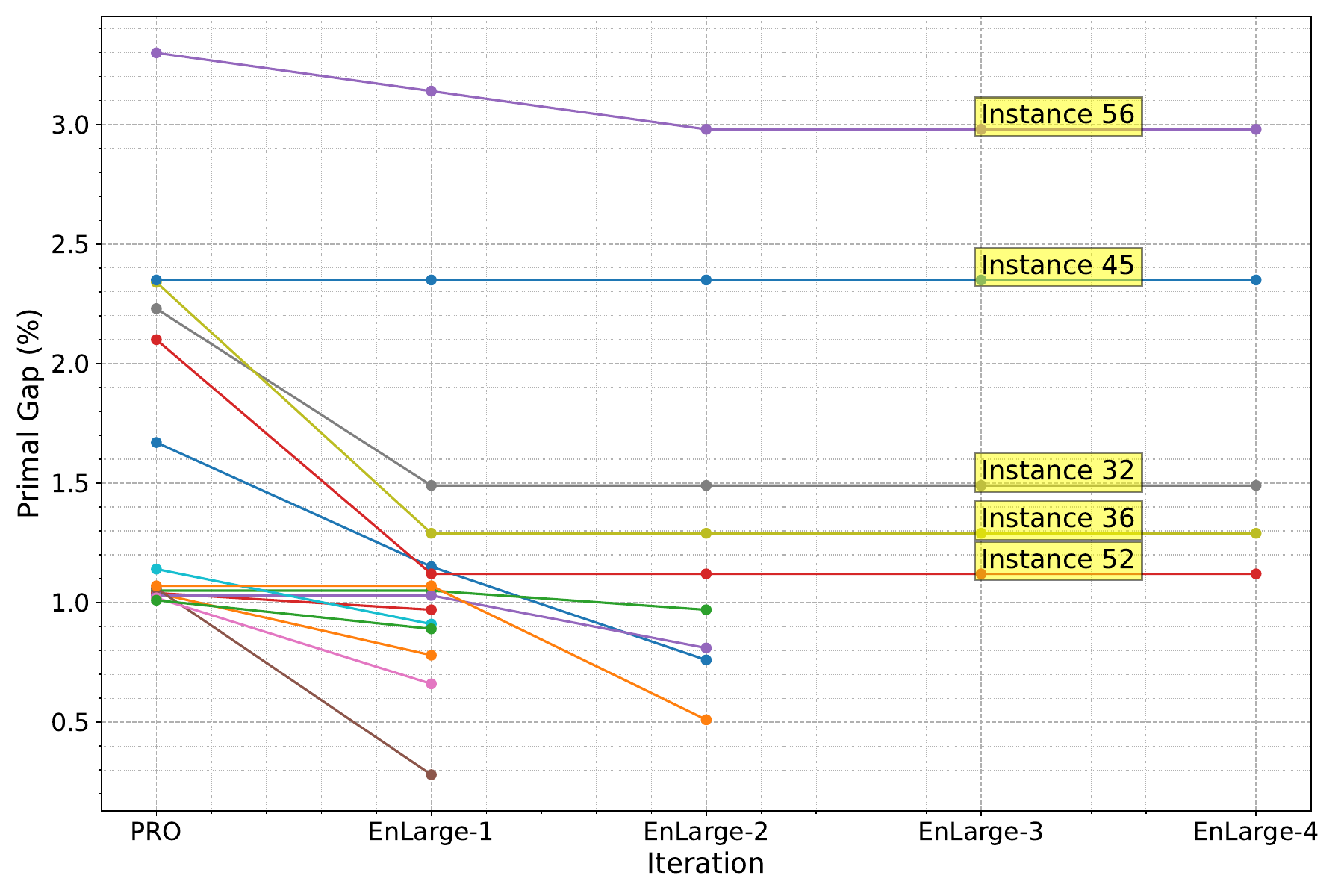}
\caption{\label{RL_gap_iterations}Primal Gap Reductions over {\sc EnLarge} Iterations.}
\end{figure}

\paragraph*{Primal Gaps}

\begin{figure}
\centering
\includegraphics[width=\linewidth]{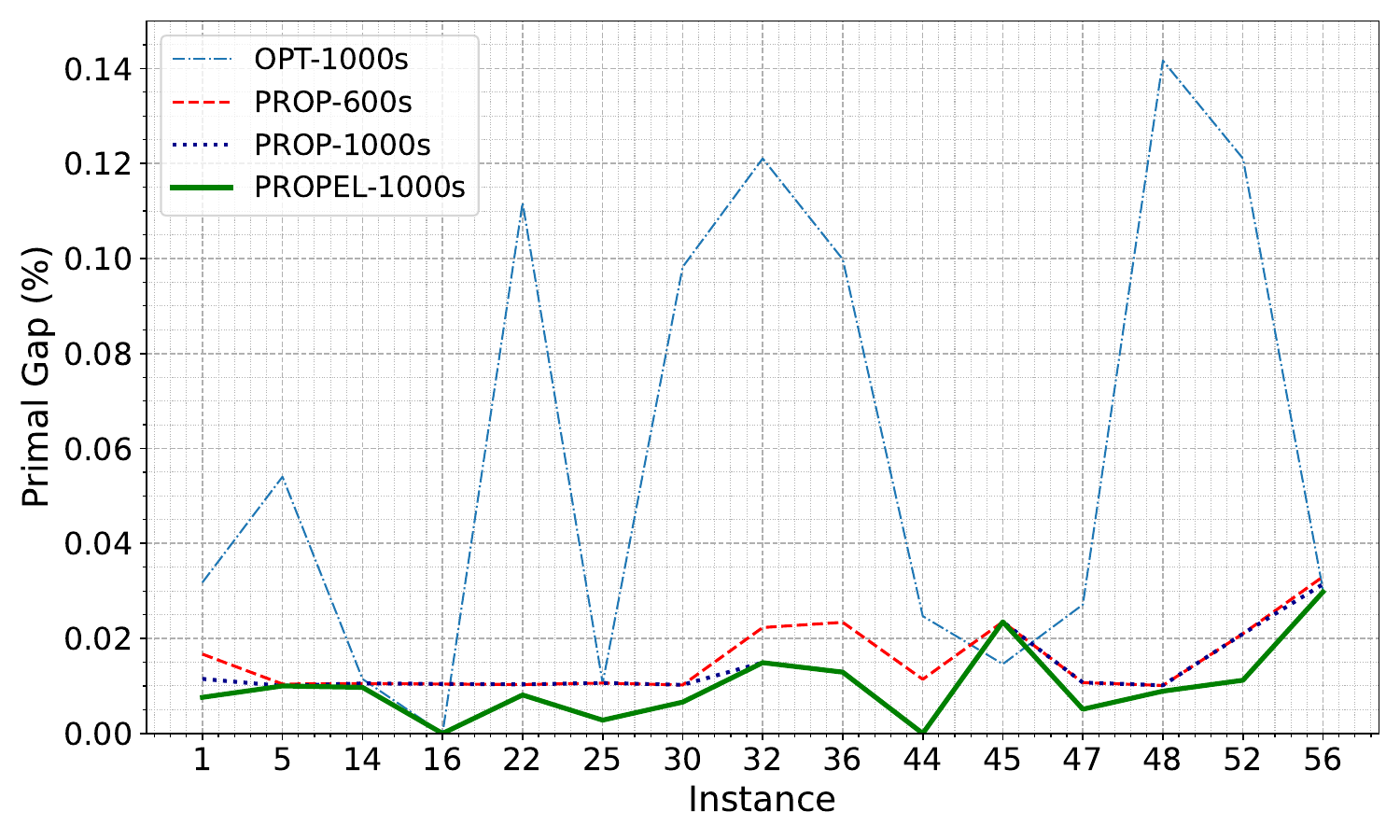}
\caption{Primal Gap Values after 1000 Seconds.}
\label{RL_Gap}
\end{figure}

Figure \ref{RL_Gap} shows that, in all but one case (instance 45),
\propel{} outperforms the other methods. Table \ref{tab:RL_gap_imp}
specifically outlines the incremental improvements in primal gaps
observed at PROP and subsequent {\sc EnLarge} iterations. The average
primal gap decreases significantly from OPT (6.35\%) to \prop{}
(1.67\%) and \propel{} (1.35\%). Even after 1000 seconds, OPT and
\prop{} exhibited primal gaps above 1\% for 37 and 14 instances,
respectively. Figure \ref{RL_gap_iterations} shows that in most cases,
the primal gap of \propel{} was reduced to 1\% within just two {\sc
EnLarge} iterations, except for Instances 32, 36, 45, 52, and 56 with
primal gaps of 2.23\%, 2.34\%, 2.35\%, 2.10\%, and 3.30\%.  {\em
Overall, these experiments demonstrate that \propel{} improves the
primal gap by a factor of up to 15.92, with an average improvement of
6.32; underscoring the effectiveness of the RL component in
reintegrating meaningful variables into the optimization.}

\begin{table}
  \centering
  \caption{Primal Gap Improvements After Each Iteration.}
    \resizebox{\columnwidth}{!}{%
    \begin{tabular}{lrrrrrr}
    \toprule
    Instance & \multicolumn{1}{l}{OPT} & \multicolumn{1}{l}{PROP} & \multicolumn{1}{l}{EnLarge-1} & \multicolumn{1}{l}{EnLarge-2} & \multicolumn{1}{l}{EnLarge-3} & \multicolumn{1}{l}{EnLarge-4} \\
    \midrule
    1     & 3.18\% & 1.67\% & 1.15\% & 0.76\% & \multicolumn{1}{r}{-} & \multicolumn{1}{r}{-} \\
    5     & 5.40\% & 1.05\% & 0.78\% & \multicolumn{1}{r}{-} & \multicolumn{1}{r}{-} & \multicolumn{1}{r}{-} \\
    14    & 1.15\% & 1.04\% & 1.05\% & 0.97\% & \multicolumn{1}{r}{-} & \multicolumn{1}{r}{-} \\
    16    & 0.00\% & 1.03\% & 0.00\% & \multicolumn{1}{r}{-} & \multicolumn{1}{r}{-} & \multicolumn{1}{r}{-} \\
    22    & 11.17\% & 1.06\% & 1.03\% & 0.81\% & \multicolumn{1}{r}{-} & \multicolumn{1}{r}{-} \\
    25    & 1.06\% & 1.01\% & 0.28\% & \multicolumn{1}{r}{-} & \multicolumn{1}{r}{-} & \multicolumn{1}{r}{-} \\
    30    & 9.82\% & 1.02\% & 0.66\% & \multicolumn{1}{r}{-} & \multicolumn{1}{r}{-} & \multicolumn{1}{r}{-} \\
    32    & 12.11\% & 2.23\% & 1.49\% & 1.49\% & 1.49\% & 1.49\% \\
    36    & 9.98\% & 2.34\% & 1.29\% & 1.29\% & 1.29\% & 1.29\% \\
    44    & 2.47\% & 1.14\% & 0.91\% & \multicolumn{1}{r}{-} & \multicolumn{1}{r}{-} &  \\
    45    & 1.46\% & 2.35\% & 2.35\% & 2.35\% & 2.35\% & 2.35\% \\
    47    & 2.71\% & 1.07\% & 1.07\% & 0.51\% & \multicolumn{1}{r}{-} & \multicolumn{1}{r}{-} \\
    48    & 14.17\% & 1.01\% & 0.89\% & \multicolumn{1}{r}{-} & \multicolumn{1}{r}{-} & \multicolumn{1}{r}{-} \\
    52    & 12.11\% & 2.10\% & 1.12\% & 1.12\% & 1.12\% & 1.12\% \\
    56    & 3.02\% & 3.30\% & 3.14\% & 2.98\% & 2.98\% & 2.98\% \\
    \bottomrule
    \end{tabular}%
    }
  \label{tab:RL_gap_imp}%
\end{table}%

\paragraph*{Runtimes}
Across all experiments, {\em \propel{} emerged as the best-performing
model, achieving termination times that are, on average, 17.5\%
shorter than OPT.} This demonstrates that \propel{} brings benefits,
not only in solution quality, but also in computational
speed. Detailed comparisons of termination times between OPT and
\propel{} are available in Table \ref{tab:cutoff}. {\em These results on
large-scale real-world instances highlight the significant benefits
of the combination of supervised and reinforcement learning at the
core of \propel{}.}

\begin{table*}
  \centering
  \caption{The total primal integral values at $t_{max}=1000$ seconds across all methods}
    \begin{tabular}{lrrrrrrrrrrrr}
    \toprule
    \multicolumn{1}{c}{\multirow{2}[4]{*}{Instasnce}} & \multirow{2}[4]{*}{} & \multicolumn{1}{c}{\multirow{2}[4]{*}{OPT-1000s}} & \multirow{2}[4]{*}{} & \multicolumn{1}{c}{\multirow{2}[4]{*}{PROP-1000s}} & \multirow{2}[4]{*}{} & \multicolumn{7}{c}{\propel{}-1000s} \\
\cmidrule{7-13}          &       &       &       &       &       & \multicolumn{1}{c}{PROP-600s} & \multicolumn{1}{c}{EnLarge-1} & \multicolumn{1}{c}{EnLarge-2} & \multicolumn{1}{c}{EnLarge3} & \multicolumn{1}{c}{EnLarge-4} &       & \multicolumn{1}{c}{Total} \\
\cmidrule{1-1}\cmidrule{3-3}\cmidrule{5-5}\cmidrule{7-11}\cmidrule{13-13}    1     &       &          146.74  &       &              83.67  &       &           75.99  & 1.44  & 0.95  &       &       &       &        78.38  \\
    5     &       &          140.13  &       &              39.33  &       &           33.18  & 0.91  &       &       &       &       &        34.09  \\
    14    &       &          124.97  &       &              33.35  &       &           31.37  & 1.05  & 1.01  &       &       &       &        33.43  \\
    16    &       &          238.78  &       &              59.68  &       &           58.79  & 1     &       &       &       &       &        59.79  \\
    22    &       &             82.89  &       &              33.93  &       &           33.79  & 0.67  &       &       &       &       &        34.46  \\
    25    &       &             87.17  &       &              33.14  &       &           31.87  & 1.03  & 0.92  &       &       &       &        33.82  \\
    30    &       &          141.07  &       &              99.07  &       &           97.68  & 0.84  &       &       &       &       &        98.52  \\
    32    &       &          235.85  &       &              95.00  &       &           86.97  & 1.86  & 1.49  & 1.49  & 1.49  &       &        93.30  \\
    36    &       &          146.94  &       &           107.23  &       &           98.29  & 1.82  & 1.29  & 1.29  & 1.29  &       &     103.98  \\
    44    &       &          160.91  &       &              86.17  &       &           81.60  & 1.03  &       &       &       &       &        82.63  \\
    45    &       &          138.12  &       &              54.69  &       &           53.80  & 2.35  & 2.35  & 2.35  & 2.35  &       &        63.20  \\
    47    &       &          138.96  &       &              34.18  &       &           33.73  & 1.07  & 0.79  &       &       &       &        35.59  \\
    48    &       &          212.01  &       &              68.38  &       &           67.73  & 0.95  &       &       &       &       &        68.68  \\
    52    &       &          191.45  &       &              77.95  &       &           61.84  & 1.61  & 1.12  & 1.12  & 1.12  &       &        66.81  \\
    56    &       &          192.20  &       &           105.90  &       &           94.24  & 3.22  & 3.06  & 3.02  & 2.98  &       &     106.52  \\
    \bottomrule
    \end{tabular}%
  \label{tab:RL_AUC}%
\end{table*}%

\begin{table}
  \centering
  \caption{Comparison of Percentage of Non-Zero Integer Variables Across Methods}
    \resizebox{\columnwidth}{!}{%
    \begin{tabular}{lrrrrr}
    \toprule
    Instasnce & \multicolumn{1}{c}{PROP} & \multicolumn{1}{c}{EnLarge-1} & \multicolumn{1}{c}{EnLarge-2} & \multicolumn{1}{c}{EnLarge-3} & \multicolumn{1}{c}{EnLarge-4} \\
    \midrule
    1     & 49.94\% & 58.28\% & 82.13\% & -     & - \\
    5     & 45.58\% & 57.22\% & -     & -     & - \\
    14    & 49.67\% & 58.02\% & 85.90\% & -     & - \\
    16    & 55.25\% & 85.90\% & -     & -     & - \\
    22    & 38.64\% & 51.55\% & 85.67\% & -     & - \\
    25    & 52.26\% & 74.27\% & -     & -     & - \\
    30    & 39.95\% & 54.29\% & -     & -     & - \\
    32    & 40.28\% & 56.23\% & 82.38\% & 92.60\% &  \\
    36    & 40.28\% & 55.57\% & 83.53\% & 92.16\% & 99.27\% \\
    44    & 38.98\% & 50.27\% & -     & -     & - \\
    45    & 42.04\% & 56.67\% & 82.13\% & -     & - \\
    47    & 37.87\% & 45.09\% & 84.47\% & -     & - \\
    48    & 53.77\% & 59.62\% & -     & -     & - \\
    52    & 39.91\% & 52.86\% & 85.90\% & 95.22\% & 98.59\% \\
    56    & 39.02\% & 50.27\% & 85.25\% & 95.58\% & 96.92\% \\
    \bottomrule
    \end{tabular}%
    }
  \label{invar}%
\end{table}%

\begin{table}
  \centering
  \caption{Comparing Runtimes between OPT and \propel{}}
    \begin{tabular}{lrr}
    \toprule
    Instance & \multicolumn{1}{l}{OPT} & \multicolumn{1}{l}{ \propel{} } \\
    \midrule

    1     &                1,000  &                     725  \\
    5     &                1,000  &                     639  \\
    14    &                     769  &                     725  \\
    16    &                     245  &                     316  \\
    22    &                1,000  &                     772  \\
    25    &                     343  &                     409  \\
    30    &                1,000  &                     645  \\
    32    &                1,000  &                1,000  \\
    36    &                1,000  &                1,000  \\
    44    &                1,000  &                     694  \\
    45    &                1,000  &                1,000  \\
    47    &                1,000  &                     795  \\
    48    &                1,000  &                     700  \\
    52    &                1,000  &                1,000  \\
    56    &                1,000  &                1,000  \\
    \bottomrule
    \end{tabular}%
  \label{tab:cutoff}%
\end{table}%

\section{Conclusion and Future Direction}
\label{section:conclusion}
This paper introduced \propel{}, a novel learning-based optimization
framework designed to enhance the computational efficiency of solving
large-scale supply chain planning optimization problems. By leveraging
a combination of supervised learning and deep reinforcement
learning, \propel{} aims at reducing the size of the search space,
thereby accelerating the finding of high-quality solutions to SCP
problems.  These problems can be formulated as MIP models which
feature both integer (non-binary) and continuous variables, and flow
balance and capacity constraints, raising fundamental challenges
for integrations of ML and optimization.
\propel{} uses supervised learning, not to predict the values of all
integer variables, but to identify those variables that are fixed to
zero in the optimal solution. It also leverages the linear
relaxation and reduced costs to improve the predictions.
\propel{} includes DRL component that selects which fixed-at-zero variables must
be relaxed to improve solution quality when the supervised learning step
does not produce a solution with the desired optimality
tolerance. \propel{} has been applied to industrial supply chain
planning optimizations with millions of variables. The computational
results show dramatic improvements in solution times and quality,
including a 60\% reduction in primal integral and an 88\% primal gap
reduction, and improvement factors of up to 13.57 and 15.92,
respectively.

To a large extent, \propel{} is a generic framework and hence would
apply to other applications that share the same characteristics,
e.g., large numbers of non-binary integer variables that are fixed at
zero. Future research will be geared towards looking for applications
beyond SCP problems, that may benefit
from \propel{}. Although \propel{} has been applied to instances with
millions of variables, scaling to even larger problems would need to
address the computational cost of solving the optimization problems
for the training instances.  It would be interesting to study
how \propel{} would apply to other solution techniques, such as large
neighborhood search.

\section*{Acknowledgement}
The authors are grateful to Kinaxis Corp. for providing data
resources. We are particularly grateful for the advice of Carsten
Jordan (Product Owner, Supply Chain Solutions), Dan Vlasie (Staff
Software Developer), Ingrid Bongartz (Kinaxis Product Manager), and
Sebastien Ouellet (Machine Learning Developer) who helped us with data
collection, answered questions, and suggested improvements to the
paper. The research is partly supported by the NSF AI Institute for
Advances in Optimization (Award 2112533).

\newpage

\appendix

\section{Evaluation Metrics and Reward Design}
\label{section:metrics}

To gauge PROPEL's efficacy, the widely adopted \textit{Primal
  Integral} metric \citep{achterberg2012rounding} is employed. This
metric assesses the average solution quality achieved within a given
time frame $t$ during MIP solving, which is the integral on $[0, t]$
of the primal gap as a function of runtime. Primal integral captures
the quality of the solutions found and the speed at which they are
found. A smaller primal gap signifies superior performance, indicating
the attainment of high-quality solutions early in the solving process.

The calculation of primal integral requires an optimal or best integer
solution. Since it may require a significant amount of time to find
$\textbf{x}^*$, the quality of obtained solutions for each instance is
benchmarked relative to the lower bound established by solving the LP
relaxation objective ($LP^*$). Accordingly, our customized primal gap
$\omega\in[0,1]$ of solution $\hat{\textbf{x}}$ is defined as:
\[
\omega(\hat{\textbf{x}}) =
\begin{cases} 
0, & \text{if } |LP^*| = |\textbf{c}^T \hat{\textbf{x}}| = 0 \\ 
1, & \text{if } LP^* \cdot \textbf{c}^T \hat{\textbf{x}} < 0 \\ 
\frac{|\textbf{c}^T \hat{\textbf{x}} - LP^*|}{\max \{|\textbf{c}^T \hat{\textbf{x}}|, |LP^*|\}}, & \text{otherwise.}
\end{cases}
\]
Then, the primal gap function $p: [o, t_{max}] \rightarrow [0, 1]$,
where $t_{max} \in \mathbb{R}_{\ge 0}$ is a limit on the solution time
of the MIP (B\&B), defined as
\[
p(t) =
\begin{cases} 
1, & \text{if no incumbent is found until point } t\\ 
\omega(\hat{\textbf{x}}(t)), & \text{with }\hat{\textbf{x}}(t) \text{ the incumbent at point t.}
\end{cases}
\]
Finally, the primal integral $\mathcal{P}(T)$ of a MIP until a point
in time $T \in [0, t_{max}]$ is defined as
\[
\mathcal{P}(T) =\sum_{i=1}^{\nu+1} p(t_{i-1}) (t_i - t_{i-1}),
\]
where $\nu$ is the number of incumbents and $t_i \in [0, T]$ for $i
\in \{1, . . . , \nu\}$ are the points in time when a new incumbent is
found, $t_0 = 0$ and $t_{\nu+1} = T$. The smaller
$\mathcal{P}(t_{max})$ is, the better the incumbent finding. As such,
the focus is on optimizing the primal integral by making better
decisions regarding whether an integer decision variable should be
eliminated from the search space.

\end{document}